\documentclass{article}

\PassOptionsToPackage{numbers,compress}{natbib}
\usepackage[main, final]{neurips_2026}
\usepackage[utf8]{inputenc} 
\usepackage[T1]{fontenc}    
\usepackage{hyperref}       
\usepackage{url}            
\usepackage{booktabs}       
\usepackage{amsfonts}       
\usepackage{nicefrac}       
\usepackage{microtype}      
\usepackage[table,xcdraw]{xcolor}
\usepackage{float}
\usepackage{amsmath,amssymb,mathtools}    
\usepackage{caption}
\usepackage{makecell}
\usepackage{tikz}
\usetikzlibrary{spy, calc}
\usepackage{soul}
\usepackage{algorithm}
\usepackage{multirow}
\usepackage{graphicx}
\usepackage{float}
\usepackage{subcaption}
\usepackage{rotating}
\usepackage{array}
\usepackage{stackengine}
\usepackage{booktabs}
\usepackage{algpseudocode}
\usepackage{cancel}
\usepackage{diagbox}
\usepackage{colortbl}
\usepackage{tabulary}
\usepackage{pgfkeys}
\usepackage{tabularray}
\usepackage{mwe}
\usepackage{enumitem}
\usepackage{wrapfig2}
\usepackage{placeins}
\usepackage{needspace}
\UseTblrLibrary{booktabs} %

\usepackage{tabularx}

\usepackage[most]{tcolorbox}
\tcbuselibrary{listings,breakable}
\usepackage{listings}

\definecolor{Gray}{gray}{0.9}
\definecolor{Red}{rgb}{0.768, 0.054, 0.054}
\definecolor{Blue}{rgb}{0.152, 0.294, 0.925}
\definecolor{URL}{rgb}{0,0.4,0.7}
\definecolor{Green}{rgb}{0,0.9,0}

\hypersetup{
    colorlinks=true,
    citecolor=teal,
    linkcolor=Red,
    urlcolor=URL,
}

\usepackage{pgfplots}
\usepackage{adjustbox}
\pgfplotsset{compat=1.18}

\definecolor{LightBlueCell}{RGB}{221,238,246}
\definecolor{LightGreenCell}{RGB}{226,243,218}

\newcommand{\bestcell}[1]{\cellcolor{LightBlueCell}{#1}}

\usepackage{tikz}
\usetikzlibrary{spy, calc}

\tikzset{
    largewindow_w/.style={white, line width=0.30mm},
    smallwindow_w/.style={white, line width=0.15mm},
    closeup_w_3/.style={
        opacity=1.0,
        height=1.35cm,
        width=1.35cm,
        connect spies,
        white
    },
}

\newcommand{\SdfZoom}[3]{%
  \begin{tikzpicture}[
      spy using outlines,
      baseline=(FigA.south west)
    ]
      \node[anchor=south west, inner sep=0] (FigA) at (0,0) {%
          \includegraphics[width=#2,clip]{#1}%
      };

    \spy [closeup_w_3, magnification=3.0,
          size=0.8cm,
          draw=red,
          line width=0.50mm,
          every spy on node/.append style={line width=0.28mm}]
      on ($(FigA)+(0.05, 0.9)$)
      in node[draw=red, line width=0.30mm, anchor=east, size=1.0cm]
          at ($(FigA.north)+(1.0, -2.0)$);

    \node[
        anchor=north, 
        font=\small
    ] at ($(FigA.south)+(0, 0.05)$) {#3}; %
  \end{tikzpicture}%
}

\newcommand{\SdfZoomDragon}[3]{%
  \begin{tikzpicture}[
      spy using outlines,
      baseline=(FigA.south west)
    ]
      \node[anchor=south west, inner sep=0] (FigA) at (0,0) {%
          \includegraphics[width=#2,clip]{#1}%
      };

    \spy [closeup_w_3, magnification=1.3,
          size=0.8cm,
          draw=red,
          line width=0.50mm,
          every spy on node/.append style={line width=0.28mm}]
      on ($(FigA)+(-0.8, 0.47)$)
      in node[draw=red, line width=0.30mm, anchor=east, size=1.0cm]
          at ($(FigA.north)+(1.15, 0.083)$);

    \node[
        anchor=north, 
        font=\small
    ] at ($(FigA.south)+(0, 0.05)$) {#3}; %
  \end{tikzpicture}%
}

\newcommand{\LegoZoom}[3]{%
  \begin{tikzpicture}[
      spy using outlines,
      baseline=(FigA.south west)
    ]
      \node[anchor=south west, inner sep=0] (FigA) at (0,0) {%
          \includegraphics[width=#2,clip]{#1}%
      };

    \spy [closeup_w_3, magnification=3.0,
          size=0.8cm,
          draw=red,
          line width=0.50mm,
          every spy on node/.append style={line width=0.28mm}]
      on ($(FigA)+(-0.3, 1.23)$)
      in node[draw=red, line width=0.30mm, anchor=east, size=1.5cm]
          at ($(FigA.north)+(1.5, -5.2)$);

    \node[
        anchor=north, 
        font=\small
    ] at ($(FigA.south)+(0, 0.05)$) {#3}; %
  \end{tikzpicture}%
}

\usepackage[percent]{overpic}
\usepackage{tikz}

\newcommand{\PSNRZoomImage}[7]{%
\begin{tikzpicture}[
    spy using outlines,
    baseline=(FigA.south west)
]
    \node[anchor=south west, inner sep=0pt, outer sep=0pt] (FigA) at (0,0)
        {\includegraphics[width=#2,clip]{#1}};

    \spy [closeup_w_3, magnification=4.0,
          size=0.8cm,
          draw=red,
          line width=0.50mm,
          every spy on node/.append style={line width=0.28mm}]
      on ($(FigA)+(#4, #5)$)
      in node[draw=red, line width=0.30mm, anchor=east, size=1.0cm]
          at ($(FigA.north)+(#6, #7)$);

    \begin{scope}
        \clip (FigA.south west) rectangle (FigA.north east);
        \node[
            anchor=south west,
            fill=black,
            text=white,
            font=\bfseries\scriptsize,
            inner sep=2pt,
            outer sep=0pt
        ] at (FigA.south west) {#3};
    \end{scope}
\end{tikzpicture}%
}

\newcommand{\AbGTZoom}[3]{%
  \begin{tikzpicture}[
      spy using outlines,
      baseline=(FigA.south west)
    ]
      \node[anchor=south west, inner sep=0] (FigA) at (0,0) {%
          \includegraphics[width=#2,clip]{#1}%
      };

    \spy [closeup_w_3, magnification=4.0,
          size=0.8cm,
          draw=red,
          line width=0.50mm,
          every spy on node/.append style={line width=0.28mm}]
      on ($(FigA)+(-0.08, 0.53)$)
      in node[draw=red, line width=0.30mm, anchor=east, size=1.0cm]
          at ($(FigA.north)+(1.195, -1.32)$);

    \node[
        anchor=north, 
        font=\small
    ] at ($(FigA.south)+(0, 0.05)$) {#3}; %
  \end{tikzpicture}%
}

\newcommand{\DinoZoom}[3]{%
  \begin{tikzpicture}[
      spy using outlines,
      baseline=(FigA.south west)
    ]
      \node[anchor=south west, inner sep=0] (FigA) at (0,0) {%
          \includegraphics[width=#2,clip]{#1}%
      };

    \spy [closeup_w_3, magnification=3.0,
          size=0.8cm,
          draw=red,
          line width=0.50mm,
          every spy on node/.append style={line width=0.28mm}]
      on ($(FigA)+(-1.4, -1.45)$)
      in node[draw=red, line width=0.30mm, anchor=east, size=1.5cm]
          at ($(FigA.north)+(1.5, -2.3)$);

    \node[
        anchor=north, 
        font=\small
    ] at ($(FigA.south)+(0, 0.05)$) {#3}; %
  \end{tikzpicture}%
}

\newcommand{\ShipZoom}[3]{%
  \begin{tikzpicture}[
      spy using outlines,
      baseline=(FigA.south west)
    ]
      \node[anchor=south west, inner sep=0] (FigA) at (0,0) {%
          \includegraphics[width=#2,clip]{#1}%
      };

    \spy [closeup_w_3, magnification=3.5,
          size=0.8cm,
          draw=red,
          line width=0.50mm,
          every spy on node/.append style={line width=0.28mm}]
      on ($(FigA)+(-0.03, -0.05)$)
      in node[draw=red, line width=0.30mm, anchor=east, size=1.5cm]
          at ($(FigA.north)+(1.6, -0.8)$);

    \node[
        anchor=north, 
        font=\small
    ] at ($(FigA.south)+(0, 0.05)$) {#3}; %
  \end{tikzpicture}%
}

\title{ProPolar: Progressive Polar Decomposition for  Implicit Neural Representations}

\author{%
  Pureum Kim \\
  Korea University \\
  \texttt{pureum\_kim@korea.ac.kr}
  \And 
  Younggeon Ryu \\
  Korea University \\
  \texttt{alphayoung8@korea.ac.kr}
  \And 
  Dongyoon Lee \\
  Korea University \\
  \texttt{dongyoon001@gmail.com}
  \AND
  Hae Beom Lee\textsuperscript{\textdagger} \\
  Korea University \\
  \texttt{haebeomlee@korea.ac.kr}
  \And 
  Kyong Hwan Jin\textsuperscript{\textdagger} \\
  Korea University \\
  \texttt{kyong\_jin@korea.ac.kr}
}

\begin{document}
\maketitle

\begingroup
\renewcommand{\thefootnote}{\textdagger}
\footnotetext{Co-corresponding authors.}
\endgroup

\vspace{-5pt}
\begin{abstract}
Implicit neural representations (INRs) capture global low-frequency structure during the early stage of training and then refine localized high-frequency details, afterwards. However, standard optimizers are agnostic to the coarse-to-fine learning behavior of INRs. Such optimizers accumulate a gradient matrix in momentum buffer and retain components misaligned with the dominant low-frequency directions in the early stage. 
We propose a progressively growing rank scheduler and a rank-aware learning-rate scheduler for enhancing a subspace-based momentum optimization, where a momentum matrix is updated within a low-rank subspace. The rank schedule expands the subspace during training, and the scheduler holds the peak learning rate through rank growth so that the added subspace contributes to fine-detail recovery.
Experiments on image fitting, single-image super-resolution (SISR), and neural radiance field optimization show that the proposed method improves convergence and peak reconstruction quality over competitive optimizers.
The largest gains appear in the high-frequency refinement stage, mirroring the coarse-to-fine progression that the rank schedule targets. A significant study on hyperparameter optimization (HPO) on NeRF further confirms that these gains persist under matched search budgets. \url{https://github.com/pur3um/ProPolar}
\end{abstract}

\section{Introduction}
\vspace{-0.05in}
 
Implicit neural representations (INRs) represent discrete signals as continuous functions parameterized by coordinate-based neural networks~\cite{chen2021learning, sitzmann2020implicit, tancik2020fourier, tancik2021learned}.
Their training exhibits the coarse-to-fine progression in which global low-frequency structure forms first and high-frequency details are refined later~\cite{hertz2021sape, rahaman2019spectral, wang2023neus2, xu2019frequency, xu2025understanding}.
Such a stage-dependent process suggests that requested capacity increases proportional to training iterations~\cite{xu2019frequency, zhang2023adalora, zhao2024galore}: low-frequency reduction in the early stage benefits from a low-rank subspace, whereas high-frequency recovery in the later stage benefits from a high-rank subspace~\cite{mahdavinia2025low, schotthofer2025geometric, wang2026taming, zhao2024galore}.
INR optimization therefore benefits from regulating the update structure throughout training.

\begin{figure*}[t]
    \centering
    \includegraphics[width=1\textwidth]{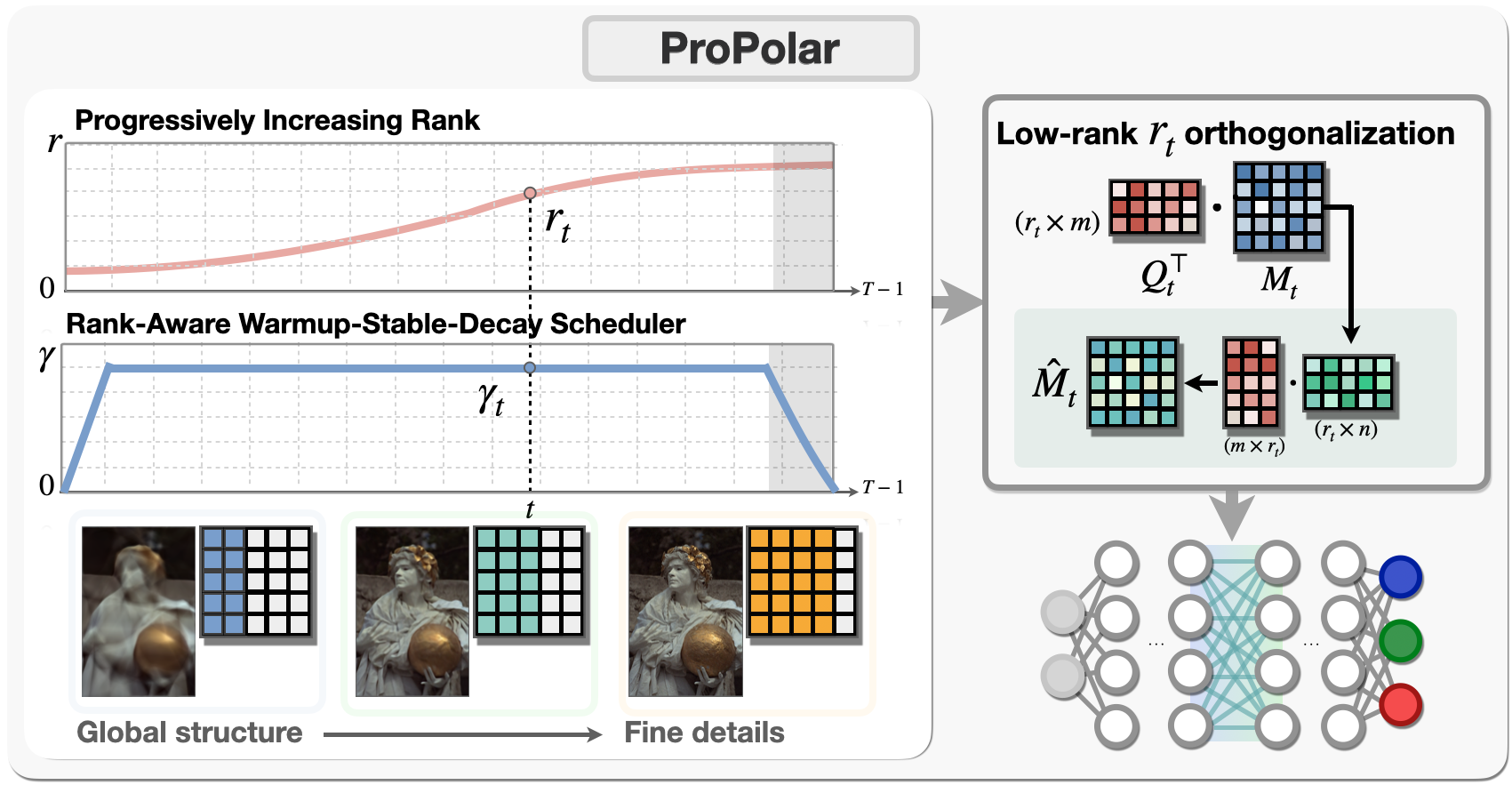}
    \caption{\small \textbf{Overview.}
ProPolar introduces a progressive active-rank schedule $r_t$ that expands the momentum subspace from dominant low-rank directions to higher-rank directions during INR training.
For the current rank budget, the orthonormal basis $Q_t\in\mathbb{R}^{m\times r_t}$ defines the active subspace used for projected polar orthogonalization.
The rank-aware WSD schedule $\gamma_t$ maintains the peak step size during rank growth, allowing newly activated directions to contribute to fine-detail recovery before decay begins.}
\label{fig:overview}
\vspace{-0.3in}
\end{figure*}

Recent matrix-aware optimizers address a regulating structure of updating by decomposing layer-wise gradients or exploiting a low-rank structure of optimizer states
~\cite{gupta2018shampoo,jordan2024muon, vyas2024soap, wang2025conda, xu2026width, zhao2024galore}. Such designs suggest the importance of regulating update-space structure during optimization. However, most optimizers retain a fixed updating representation throughout training. Full-rank optimizers use the same full-dimensional  representation throughout training~\cite{mahdavinia2025low, wang2026taming, yu2025altlora, zhao2024galore}. A momentum matrix accumulates a full-sized gradient, including components misaligned with dominant updates during the early stage. Fixed-rank optimizers assign capacity to less informative directions during the early stage and constrain the update subspace during the later detail-refinement stage~\cite{robert2024ldadam, zhao2024galore}. 
In contrast, the coarse-to-fine behavior of INR training suggests that a fixed-dimensional update representation is misaligned with stage-dependent reconstruction demands~\cite{rahaman2019spectral, schotthofer2025dynamical, xu2019frequency}. 
The alignment issue motivates an optimizer with a progressively increasing rank that updates parameters in the original parameter space.
Prior matrix-aware optimizers exploit compact update subspaces 
through low-rank projection or orthogonalization, but they are not 
designed for the coarse-to-fine training dynamics of 
INRs~\cite{he2025low, zhao2024galore}.

In this paper, we propose ProPolar, an optimizer that projects matrix-valued momentum onto a progressively expanding subspace to match the coarse-to-fine dynamics of INR training.
As illustrated in Figure~\ref{fig:overview}, early updates concentrate on a small set of dominant directions associated with low-frequency structures, while the active subspace broadens later in training as fine-detail recovery proceeds.
To preserve the contribution of these later directions, we pair the rank schedule with a learning-rate schedule that maintains the peak step size during rank growth. This combination aligns the optimizer state with the coarse-to-fine learning pattern of INRs~\cite{rezaeian2025sl2a, xu2019frequency}.

Experiments on image fitting, single-image super-resolution (SISR), and neural radiance field optimization show that the proposed method improves convergence and reconstruction quality over prior alternatives.
Since INR training can be sensitive to optimization hyperparameters, we further conduct a unified tree-structured parzen estimator (TPE)-based HPO comparison on NeRF, where Adam, Muon, and the proposed optimizer are tuned under the same search budget~\cite{akiba2019optuna, bergstra2011algorithms, choi2019empirical}.
The proposed optimizer remains the top performer under this tuned comparison, indicating that its gains are not merely due to hyperparameter mismatch.

Our contributions are summarized as follows:
\begin{itemize}[leftmargin=*, itemsep=0.1em, topsep=0em]
\item We analyze INR optimization through stage-dependent update, identifying the misalignment between fixed-rank representations and the progressive increase in reconstruction complexity as an obstacle.
\item We propose an adaptive-rank momentum optimizer that represents momentum matrices in progressively increasing subspaces. 
In the early training phase, our optimizer concentrates on low-rank polar decomposed components and adds finer update directions as training progresses.
\item We show consistent reconstruction improvements across image fitting, single-image super-resolution, and neural radiance field optimization, and further verify that the gains persist under a unified TPE-based HPO comparison in NeRF experiments.
\end{itemize}

\section{Related Work}
\label{sec:related}
\vspace{-6pt}
\paragraph{Implicit Neural Representations.}
Implicit neural representations (INRs) parameterize discrete signals as continuous coordinate-to-value functions~\cite{sitzmann2020implicit}. Standard coordinate-based MLPs exhibit spectral bias and learn low-frequency components earlier than high-frequency details~\cite{lazzari2023understanding}. Fourier features and positional encodings mitigate this bias and enable high-frequency learning~\cite{tancik2020fourier}. Periodic activations such as SIREN improve signal reconstruction and derivative fidelity~\cite{sitzmann2020implicit}. FINER develops this line by introducing variable-periodic activations~\cite{liu2024finer}. Several INR methods exploit the coarse-to-fine annealing behavior. SAPE progressively unmasks higher-frequency components over time and space~\cite{hertz2021sape}. PINs learn multi-scale implicit representations with progressive levels of detail~\cite{landgraf2022pins}. BARF applies a similar strategy to NeRF optimization and registration~\cite{lin2021barf}. These results indicate that frequency content and reconstruction detail are not treated uniformly throughout training, which motivates optimizer designs that account for stage-dependent fitting dynamics.

\vspace{-6pt}
\paragraph{INR Optimization and Matrix-Aware Optimizers.}
Optimization affects INR reconstruction quality beyond representation design. Training-side methods address spectral bias through Fourier reparameterization or gradient adjustment~\cite{shi2024inductive, shi2024improved}. Rank-based INR studies relate reconstruction fidelity to stable rank, rank collapse, and rank restoration~\cite{mcginnis2025optimizing, zheng2026inlet}. These results suggest that update-space structure and rank-related dynamics are important considerations in high-fidelity INR fitting.
Matrix-aware optimizers exploit the structure of matrix-valued parameters rather than treating updates as unstructured vectors. Muon applies orthogonalized updates to 2D hidden-layer parameters and Adam-style updates to non-matrix parameters~\cite{jordan2024muon}. Related low-rank and low-dimensional methods use projected gradients, compressed optimizer states, or factorized momentum to reduce optimizer-state cost while retaining full-parameter updates~\cite{hao2024flora, mahdavinia2025low, robert2024ldadam, zhao2024galore}. Prior work has shown that gradient-based training tends to evolve in a low-dimensional subspace of the full parameter space~\cite{gur2018gradient, li2018measuring}.
We specialize this observation to INR optimization by representing 
matrix-valued momentum in a subspace whose rank progressively 
increases, paired with a learning-rate schedule that preserves the 
step size during rank growth.
\vspace{-4pt}

\section{Methodology}
\label{sec:method}

\vspace{-3pt}
\subsection{INR Optimization Setup}
\label{subsec:inr_opt_geometry}
\vspace{-5pt}

We consider an $L$-layer coordinate MLP 
$f_\theta : \mathcal{X} \to \mathcal{Y}$ trained by minimizing
\vspace{-12pt}

\begin{equation}
    \mathcal{L}(\theta)
    = \frac{1}{N}\sum_{i=1}^{N}\|f_\theta(x_i) - y_i\|_2^2,
    \label{eq:objective}
\end{equation}
\vspace{-7pt}

where $\theta$ denotes the trainable parameters of $f_\theta$ and 
$\{(x_i, y_i)\}_{i=1}^{N}$ are coordinate-value observations of the target signal. %
Let $\mathcal{H}$ denote the hidden-layer weight matrices of $f_\theta$, 
where each $W \in \mathcal{H}$ has shape $m \times n$.
Throughout the methodology, the terms \emph{gradient matrix} and 
\emph{momentum matrix} refer exclusively to the gradient and momentum associated with hidden-layer weight matrices.
At each iteration $t$, the optimizer maintains a Nesterov momentum 
matrix $M_t \in \mathbb{R}^{m \times n}$ for each 
$W \in \mathcal{H}$.

Section~\ref{sec:rank_analysis} shows that the spectra of $M_t$ and updated momentum matrix retain a dominant low-rank structure throughout INR training, while their effective rank grows over training.
This stage-dependent spectral behavior matches the coarse-to-fine learning pattern of INRs~\cite{rahaman2019spectral, xu2019frequency}. 
\vspace{-7pt}
\paragraph{Momentum Orthogonalization.}
Rank collapse in INR weight spectra reduces reconstruction fidelity~\cite{mcginnis2025optimizing, zheng2026inlet}. 
Matrix-aware optimizers mitigate rank collapse by orthogonalizing updates from $M_t$, yet full-space orthogonalization amplifies tail singular directions misaligned with the dominant update subspace~\cite{jordan2024muon, mcginnis2025optimizing, pang2026htmuon}.

\vspace{-6pt}
\subsection{Progressive Polar Decomposition}
\label{subsec:lr_polar}
\vspace{-5pt}

Orthogonalizing a matrix—computing its nearest semi-orthogonal matrix—yields the steepest-descent direction under the spectral norm~\cite{doi:10.1137/1.9780898717778, jordan2024muon}. Applied to the momentum matrix $M_t$, however, full-rank orthogonalization amplifies every singular vector equally, including uninformative directions~\cite{pang2026htmuon, zhao2024galore}. In INR training, the spectrum of a momentum term remains approximately low-rank throughout most of training (Section~\ref{sec:rank_analysis}), and the tail singular vectors are misaligned with the dominant  subspace (Section~\ref{sec:tail_analysis}). %
Inspired by \cite{he2025low},
we project $M_t$ onto its progressively increasing low-rank ($r_t$) subspace using a Gaussian random matrix
\(\Omega_t\in\mathbb R^{n\times r_t}\) and obtain an orthonormal column basis as follows:
\begin{equation}
    Q_t = \mathrm{qf}(M_t\Omega_t) \in \mathbb{R}^{m \times r_t}
\end{equation}

where \(\mathrm{qf}(\cdot)\) returns the \(Q\)-factor of a thin QR decomposition. Standard randomized-SVD analysis guarantees that
\(Q_tQ_t^{\top}M_t\) approximates the best rank-\(r_t\)
truncation of \(M_t\)~\cite{halko2011finding}. Let \(\mathcal{P}(\cdot)\) denote the orthogonal/polar factor of the polar decomposition which is approximated by 5-step Newton--Schulz iterations \cite{jordan2024muon}. The proposed update is calculated through a  \(r_t \times n\) reduced matrix:
\begin{equation}
    \hat{M}_t
    = Q_t \mathcal{P}(Q_t^\top M_t),
    \qquad
    \hat{M}_t \in \mathbb{R}^{m \times n}.
    \label{eq:polar_proj}
\end{equation}
The details about progressive polar decomposition based on the low-rank subspace orthogonalization \cite{he2025low} are described in Algorithm \ref{alg:ppd}.

\subsection{Progressively Increasing Rank}
\label{subsec:rank_growth}

We run optimization for $T$ iterations, indexed by $t \in \{0,\ldots,T-1\}$.
For each hidden-layer matrix $W\in\mathcal H$, let
$d=\min(m,n)$. We choose $T_0$, $T_g$, $r^{\max}$, and an energy threshold $\rho$ such that
\[
0<T_0<T,\qquad
2\le T_{\rm g}\le T-T_0,\qquad
1\le r^{\max}\le d,\qquad
\rho\in(0,1].
\]
A static rank in~\cite{jordan2024muon} is efficient during early-phase training but becomes restrictive when INRs represent fine details.
Conversely, using a large rank from the beginning can include low-energy
tail singular directions in the polar decomposition~\cite{pang2026htmuon}.
We therefore estimate a data-dependent starting rank during a short
initialization phase and then increase the rank by a cosine schedule.

\paragraph{Initialization phase ($0 \le t < T_0$).}
During the first $T_0$ iterations, we choose the rank $r_t$ from the singular values of the momentum matrix $M_t$. Let
$\sigma_{t,1}\ge\sigma_{t,2}\ge\cdots\ge\sigma_{t,d}$ denote the singular
values of $M_t$. For $M_t\neq 0$ and $k\in\{1,\ldots,d\}$, define
\begin{equation}
E_t(k)
=
\frac{\sum_{i=1}^{k}\sigma_{t,i}^2}
     {\sum_{i=1}^{d}\sigma_{t,i}^2},
\qquad
r_t
=
\min
\left\{
r^{\max},
\min\left\{
k\in\{1,\ldots,d\}: E_t(k)\ge \rho
\right\}
\right\}.
\label{eq:init}
\end{equation}
If $M_t=0$, we set $r_t=1$.
At the end of the initialization phase, we set
$r^{\rm init}=\left\lceil\operatorname{median}(r_0,\ldots,r_{T_0-1})\right\rceil$
as the initial rank for the cosine phase.

\paragraph{Cosine growth phase ($T_0 \le t < T$).}
After initialization, the rank grows from $r^{\rm init}$ to $r^{\max}$
over a rank-growth length $T_{\rm g}$. For $t\ge T_0$, the rank used in the projected polar decomposition is
\begin{equation}
r_t
=
\left\lceil
r^{\rm init}
+
(r^{\max}-r^{\rm init})
\frac{1-\cos(\pi s_t)}{2}
\right\rceil ,
\qquad
s_t
=
\min\left\{
1,\,
\frac{t-T_0}{T_{\rm g}-1}
\right\}.
\label{eq:rcos}
\end{equation}
Thus $r_t=r^{\rm init}$ at $t=T_0$, reaches $r^{\max}$ at
$t=T_0+T_g-1$, and remains at $r^{\max}$ thereafter. This schedule biases early training toward dominant directions and gradually admits higher-rank components for fine-detail recovery.
\vspace{-5pt}

\subsection{Rank-Aware Warmup-Stable-Decay Scheduler}
\label{subsec:rawsd}
\vspace{-3pt}

The increasing-rank schedule in Eq.~\eqref{eq:rcos} admits additional subspace directions over training. If learning-rate decay begins before the rank reaches its target, these later directions are updated with smaller steps and can contribute less to fine-detail recovery. The scheduler acts only through the learning-rate multiplier; the rank trajectory itself is determined by Eq.~\eqref{eq:rcos}.

Let $\tau_{\rm sat}=\min\{t:r_t=r^{\max}\}$ denote the first iteration at which the rank reaches its target. With layer-wise rank states, $\tau_{\rm sat}$ is the first iteration at which all tracked hidden-layer groups are saturated. We set the decay-start and
decay-end iterations to
\begin{equation}
T_d
=
\min\left\{
T-1,\,
\min\{t\ge T_0 : r_t=r^{\max}\}+\Delta_{\rm sat} 
\right\}.
\label{eq:rawsd_start}
\end{equation}
Let $\gamma_{\min}\in[0,1]$ be the minimum learning-rate ratio and
$\Delta_{\rm sat}\ge0$ be an optional delay after rank saturation. 
The learning-rate multiplier $\gamma_t\in[0,1]$ is defined by Eq.~\eqref{eq:rawsd}

\begin{equation}
\gamma_t
=
\begin{cases}
\dfrac{t+1}{T_0},
& 0\le t<T_0,\\[7pt]
1,
& T_0\le t<T_d,\\[7pt]
\gamma_{\min}
+
(1-\gamma_{\min})
\dfrac{T-1-t}{\max\{1,T-1-T_d\}},
& T_d\le t<T .
\end{cases}
\label{eq:rawsd}
\end{equation}

The same multiplier is applied to each parameter group $g$: $\eta_{t,g}=\gamma_t\eta^{\max}_g .$
By maintaining the peak learning rate until rank saturation, rank-aware WSD prevents
newly admitted components in the projected polar decomposition from being
immediately suppressed by learning-rate decay. The final linear decay is
applied only after the rank has saturated at $r^{\max}$.
\vspace{-3pt}
\begin{figure}[H]
\centering
\begin{minipage}[t]{0.58\textwidth}
\vspace{-8pt}
\begin{algorithm}[H]
\small
\caption{Progressive $r_t$ Polar Decomposition}
\label{alg:ppd}
\begin{algorithmic}[1]
\Require base lr $\eta_{\max}$, momentum coeff.\ $\beta$, weight decay $\lambda$;
\Statex initial rank $r_0=1$,
         lr ratio $\gamma_t$ (Eq.~\eqref{eq:rawsd})
\State $\eta_t \gets \gamma_t\,\eta_{\max}$
    \Comment{rank-aware Warmup-Stable-Decay}
\State Compute $\nabla_W \mathcal{L}_t$
\For{each $l \in \mathcal{H}$}
    \Statex \quad\emph{(layer index suppressed below)}
    \State $M_t = \beta M_{t-1} + \nabla_{W}\mathcal L_t \left(W_{t-1}-\beta M_{t-1}\right)$
    \Statex \Comment{Update momentum matrix}
    \State $r_t$ calculation from $M_t$ (Eqs.~\eqref{eq:init}--\eqref{eq:rcos})
    \State $\Omega_{t} \sim \mathcal{N}(0,1)^{n \times r_{t}}$
    \State $Q_{t} \gets \mathrm{qf}(M_{t}\,\Omega_{t})$
        \Comment{thin QR}
    \State $\hat{M}_{t} \gets Q_{t}\,\mathcal{P}(Q_{t}^{\top} M_{t})$
        \Comment{Eq.~\eqref{eq:polar_proj}}
    \State $W_{t+1} \gets (1-\eta_t\lambda)\,W_{t} - \eta_t\,\hat{M}_{t}$
\EndFor
\State Update non-matrix parameters ($W \notin \mathcal{H}$) with Adam
\end{algorithmic}
\end{algorithm}
\end{minipage}
\hfill
\begin{minipage}[t]{0.39\linewidth}
\vspace{5pt}
\captionof{table}{\small \textbf{Inputs to Algorithm~\ref{alg:ppd}.} Symbols defined
in Section~\ref{sec:method} ($\eta_\text{Adam}, \beta, r_t,
\gamma_t$) follow that section's definitions, while
$(\beta_1, \beta_2)$ are the first- and second-moment EMA
coefficients of the auxiliary Adam introduced only for the
non-matrix update on Line~11.}
\label{tab:hyperparams}

\centering
\renewcommand{\arraystretch}{1.465}
\setlength{\tabcolsep}{2.9pt}
\small
\definecolor{lightgray}{gray}{0.85}
\begin{tabular}{p{1.1cm}!{\color{lightgray}\vrule}c!{\color{lightgray}\vrule}c!{\color{lightgray}\vrule}c}
\arrayrulecolor{black}\hline\hline
\textbf{Task}
& $\eta_{\text{Adam}}$
& $\beta$
& $(\beta_1, \beta_2)$ \\
\hline\hline
Image
& \multirow{3}{*}{$1\mathrm{e}{-4}$}
& \multirow{4}{*}{$0.95$}
& \multirow{3}{*}{$(0.9,~0.999)$} \\
\arrayrulecolor{lightgray}\cline{1-1}
SDF  & & & \\
\arrayrulecolor{lightgray}\cline{1-1}
SISR & & & \\
\arrayrulecolor{black}\cline{1-2}\cline{4-4}
NeRF
& $4\mathrm{e}{-3}$
&
& $(0.9,~0.95)$ \\
\arrayrulecolor{black}\hline\hline
\end{tabular}
\end{minipage}
\end{figure}

\section{Experiments}
\label{sec:experiments}
\vspace{-7pt}

In this section, we evaluate the proposed optimizer across INR reconstruction tasks and analyze the effect of rank-adaptive projected polar decomposition.
The experiments assess fine-scale structure recovery and reconstruction quality.
Representative INR tasks cover direct signal fitting and inverse problems.
Each experiment compares the proposed method with task-specific baselines under consistent architectural and training settings.
Table~\ref{tab:hyperparams} provides the hyperparameter values used in all experiments.
Unless otherwise stated, all experiments use a single NVIDIA GeForce RTX 3090 GPU, and table captions specify the hardware for experiments on other GPUs. The Appendix provides additional experiments, schedule hyperparameters, HPO details, and the cost reduction from basis reuse.
\vspace{-0.2in}
\begin{wrapfigure}[24]{r}{0.50\textwidth}
\vspace{0.33in}
\centering
\captionof{table}{\small\textbf{Overfitting performance on Kodak24.}
Quantitative comparison on ReLU MLP and SIREN. Each entry reports the mean
over 24 images. Bold values indicate the best result.}
\vspace{-0.07in}
\label{tab:overtable}

\def\TableColSep{4pt}      %
\def\TableRowSep{1pt}      %
\small
\begin{tblr}{
    width   = \linewidth,
    colspec = {X[2.5,l] *{6}{X[1,c]}},   %
    colsep  = \TableColSep,
    rowsep  = \TableRowSep,
    cells   = {valign=m},
    row{1-2} = {rowsep=0.5pt},          %
    cell{1}{1} = {r=2}{l},
    cell{1}{2} = {c=2}{c},
    cell{1}{4} = {c=2}{c},
    cell{1}{6} = {c=2}{c},
}
\toprule
\textbf{Model}
  & \scriptsize \textbf{PSNR} $\uparrow$ &
  & \scriptsize \textbf{SSIM} $\uparrow$ &
  & \scriptsize \textbf{LPIPS} $\downarrow$ & \\
\cmidrule(lr){2-3}\cmidrule(lr){4-5}\cmidrule(lr){6-7}
  & \scriptsize Adam & \scriptsize \textbf{Ours} & \scriptsize Adam & \scriptsize \textbf{Ours} & \scriptsize Adam & \scriptsize \textbf{Ours} \\
\midrule
\scriptsize ReLU MLP & \scriptsize 22.55 & \scriptsize \textbf{28.70} & \scriptsize 0.5404 & \scriptsize \textbf{0.7855} & \scriptsize 0.5835 & \scriptsize \textbf{0.3417} \\
\scriptsize SIREN~\cite{sitzmann2020implicit}
         & \scriptsize 33.67 & \scriptsize \textbf{35.05} & \scriptsize 0.9104 & \scriptsize \textbf{0.9320} & \scriptsize 0.0754 & \scriptsize \textbf{0.0751} \\
\bottomrule
\end{tblr}

\vspace{0.03in}
\def\TableColSep{2pt}
\def\TableRowSep{1.5pt}

\newlength{\QualImgW}\setlength{\QualImgW}{0.42\linewidth}
\newlength{\RowLabelW}\setlength{\RowLabelW}{2mm}

\newsavebox{\ReluAdamBox}\newsavebox{\ReluOursBox}
\newsavebox{\SirenAdamBox}\newsavebox{\SirenOursBox}

\sbox{\ReluAdamBox}{\includegraphics[width=\QualImgW]{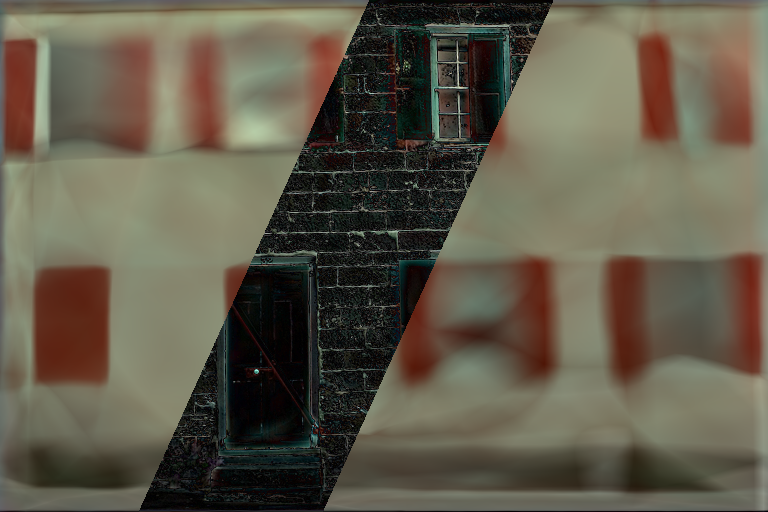}}
\sbox{\ReluOursBox}{\includegraphics[width=\QualImgW]{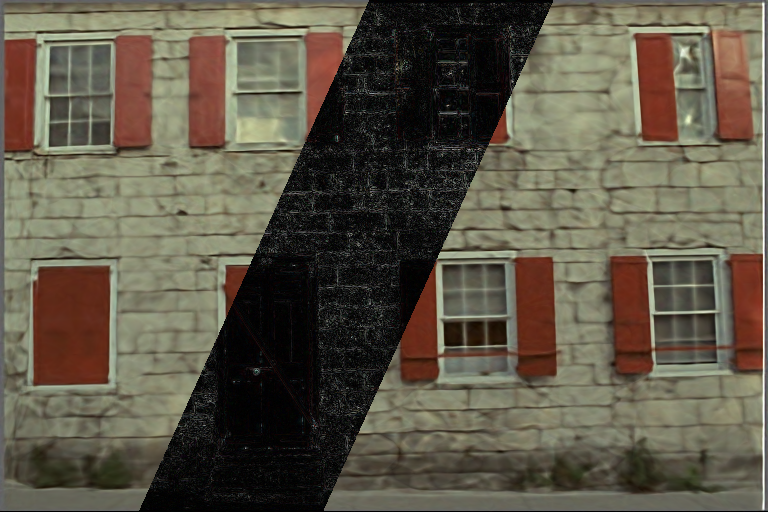}}
\sbox{\SirenAdamBox}{\includegraphics[width=\QualImgW]{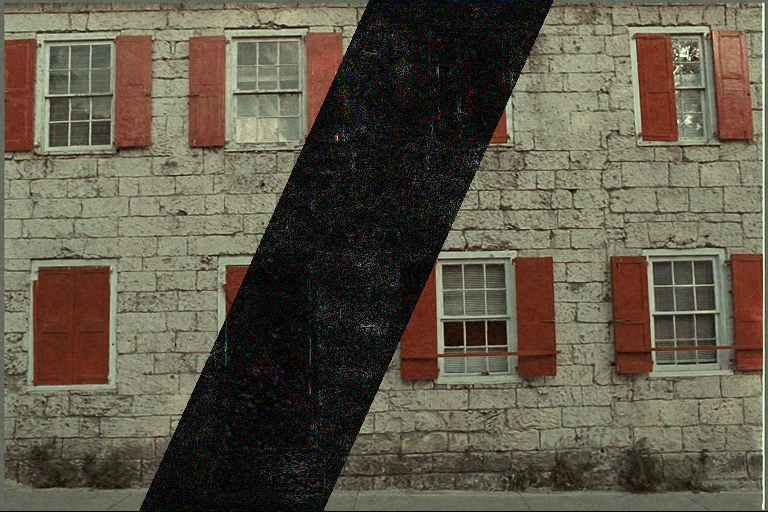}}
\sbox{\SirenOursBox}{\includegraphics[width=\QualImgW]{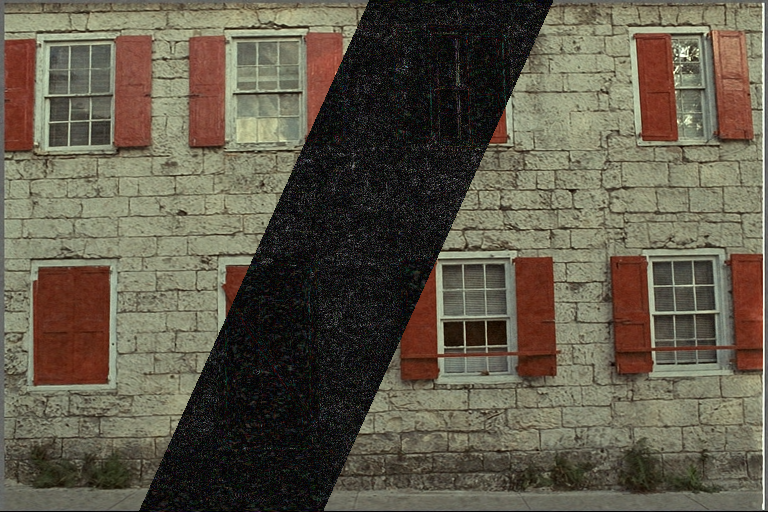}}

\newlength{\ReluRowH}\newlength{\SirenRowH}
\setlength{\ReluRowH}{\dimexpr\ht\ReluAdamBox+\dp\ReluAdamBox\relax}
\setlength{\SirenRowH}{\dimexpr\ht\SirenAdamBox+\dp\SirenAdamBox\relax}

\newcommand{\RowLabelBox}[2]{%
    \parbox[b][#1][c]{\RowLabelW}{\centering\rotatebox[origin=c]{90}{\scriptsize\strut #2}}%
}

\begin{tblr}{
    colspec = {Q[c,wd=\RowLabelW] Q[c,wd=\QualImgW] Q[c,wd=\QualImgW]},
    colsep   = 1pt,
    rowsep   = 0pt,
    cells    = {valign=m, halign=c},
    row{1}   = {font=\scriptsize},
    row{2,3} = {abovesep=0pt, belowsep=0pt},
}
     & Adam & Ours \\
    \RowLabelBox{\ReluRowH}{ReLU MLP} & \usebox{\ReluAdamBox} & \usebox{\ReluOursBox} \\
    \RowLabelBox{\SirenRowH}{Siren}   & \usebox{\SirenAdamBox} & \usebox{\SirenOursBox}
\end{tblr}

\vspace{-0.05in}
\captionof{figure}{\small\textbf{Qualitative comparison on image overfitting.}
Reconstructions from Adam and the proposed optimizer for ReLU MLP and SIREN.}
\vspace{-0.1in}
\label{fig:overimage}
\end{wrapfigure}

\subsection{Signal Representation}
\paragraph{Overfitting of Image Signals.}
We evaluate image representation performance on the 24-image Kodak dataset~\cite{kodak}.
Table~\ref{tab:overtable} reports quantitative results.
Figure~\ref{fig:overimage} shows representative reconstructions with black diagonal 
bands denoting pixel-wise absolute differences from the ground-truth images.
The proposed optimizer improves reconstruction quality across all evaluated architectures.
The results support the proposed method as an effective update rule for INR fitting.
Appendix~\ref{tab:a_app_image_overfit} extends this comparison to WIRE~\cite{saragadam2023wirewaveletimplicitneural} and FINER~\cite{liu2024finer}.

\begin{figure}[h]
\hfill
\begin{subfigure}{0.242\linewidth}
    \centering
    \SdfZoom{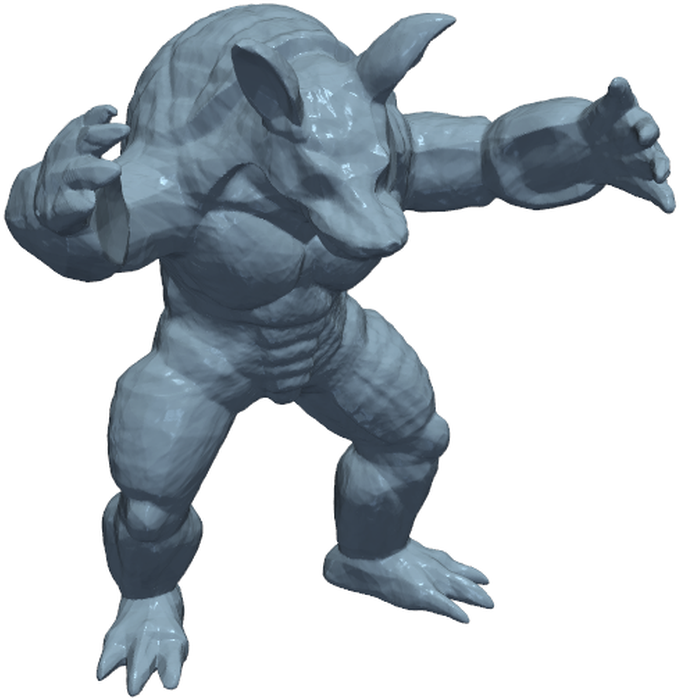}{\linewidth}{ReLU MLP w/ Adam}
\end{subfigure}
\hfill
\begin{subfigure}{0.242\linewidth}
    \centering
    \SdfZoom{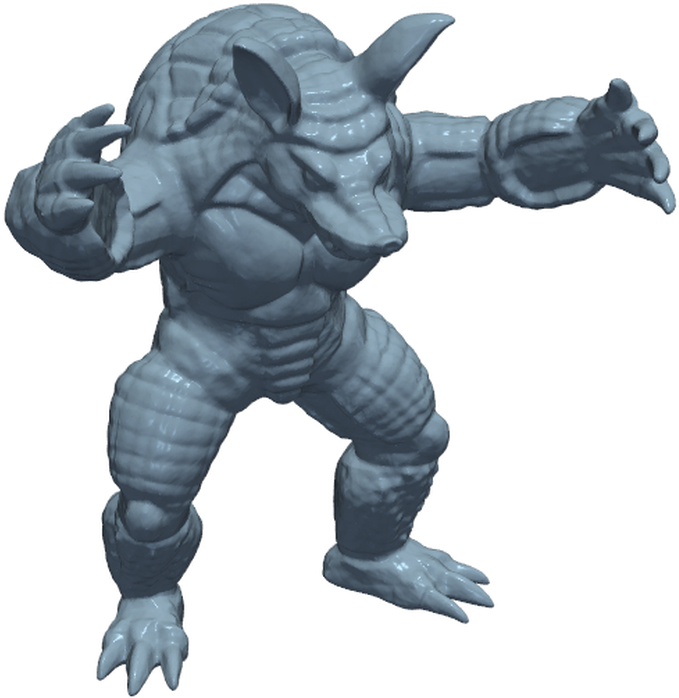}{\linewidth}{ReLU MLP w/ Ours }
\end{subfigure}
\hfill
\begin{subfigure}{0.242\linewidth}
    \centering
    \SdfZoomDragon{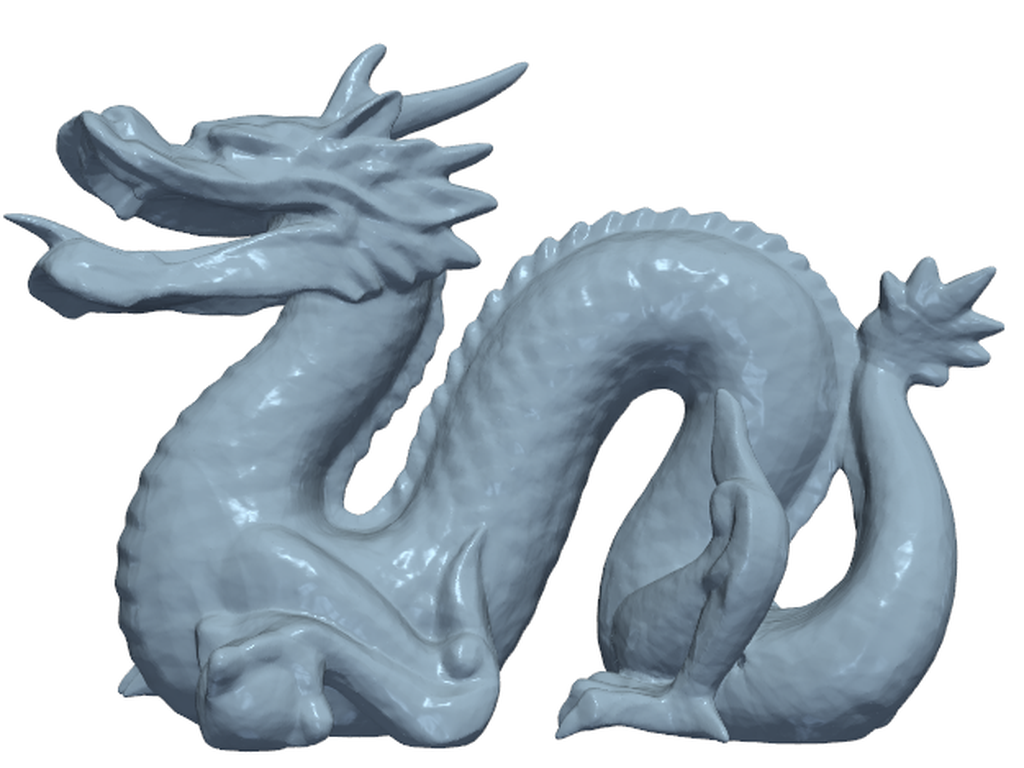}{\linewidth}{ReLU MLP w/ Adam }
\end{subfigure}
\hfill
\begin{subfigure}{0.242\linewidth}
    \centering
    \SdfZoomDragon{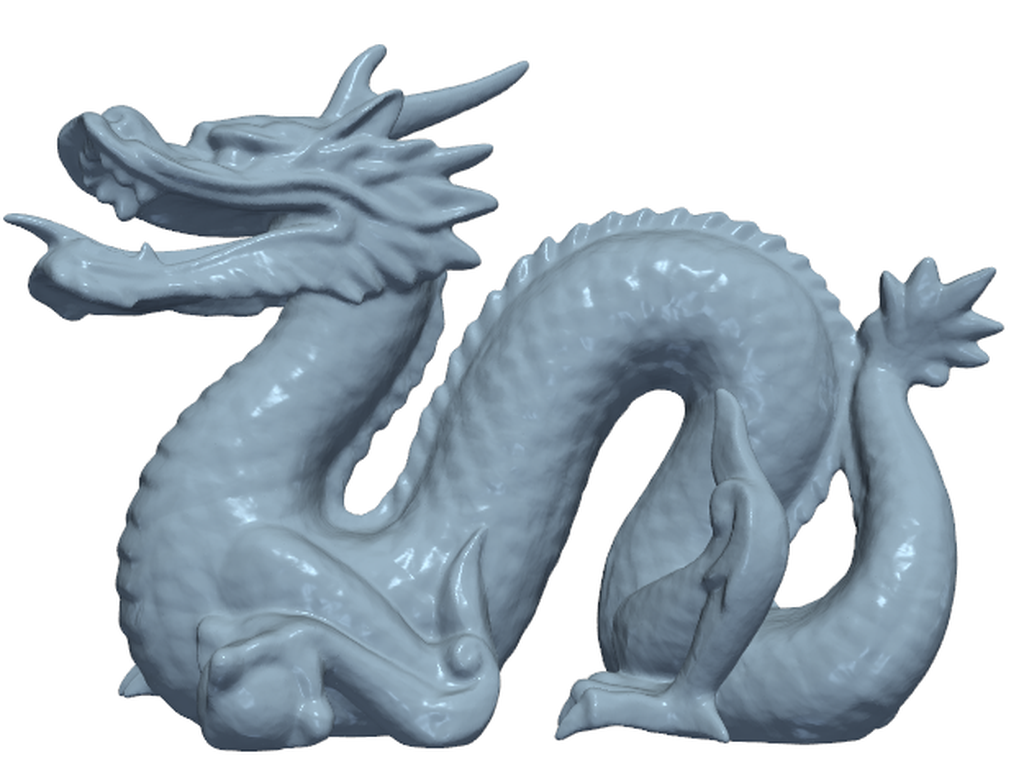}{\linewidth}{ReLU MLP w/ Ours}
\end{subfigure}
\hfill
\vspace{-0.05in}
\caption{\small\textbf{Qualitative comparison for SDF overfitting.}
Geometry reconstructions and zoomed regions on Armadillo and Dragon.} 
 \label{fig:sdf}
\end{figure}

\paragraph{Signed Distance Field Overfitting.}
We evaluate neural signed distance field (SDF) fitting.
SDF fitting requires consistency among the zero-level surface, the distance field, and spatial gradients. Following the public baseline~\cite{sitzmann2020implicit}, we use a ReLU MLP on the Armadillo and Dragon models from the Stanford 3D Scanning Repository.
Figure~\ref{fig:sdf} shows representative geometry reconstructions.
The proposed optimizer captures finer surface details than Adam, as highlighted in the zoomed regions.

\subsection{Inverse Problem}
\paragraph{Single Image Super-Resolution.}

\begin{wrapfigure}{r}{0.42\textwidth}
\vspace{-10pt}
\centering
\small
\captionof{table}{\small\textbf{Quantitative comparison.}
The best-performing optimizer for each architecture is highlighted in bold.}
\label{tab:sisr_quan}
\vspace{-2pt}

\renewcommand{\arraystretch}{0.985}%
\setlength{\tabcolsep}{2pt}%
\setlength{\aboverulesep}{0.15ex}%
\setlength{\belowrulesep}{0.15ex}%
\begin{tabular}{@{}lccc@{}}                %
\toprule[0.8pt]
\multirow{2}{*}{\textbf{Model}}
    & \multicolumn{3}{c}{\textbf{PSNR} $\uparrow$} \\
\cmidrule(lr){2-4}
& Adam~\cite{kingma2014adam} & Muon~\cite{jordan2024muon} & \bestcell{\textbf{Ours}} \\
\midrule
ReLU MLP & 19.65 & \underline{22.01} & \bestcell{\textbf{23.47}} \\
SIREN    & 24.32 & \textbf{24.73}    & \bestcell{\textbf{24.73}} \\
WIRE     & 23.97 & \underline{24.62} & \bestcell{\textbf{24.68}} \\
FINER    & \underline{25.23} & 25.19 & \bestcell{\textbf{25.45}} \\
\bottomrule[0.8pt]
\end{tabular}
\vspace{-3pt}
\end{wrapfigure}

We evaluate single-image super-resolution (SISR) on the DIV2K
dataset~\cite{Timofte_2017_CVPR_Workshops} with an upsampling factor of 4.
Table~\ref{tab:sisr_quan} reports quantitative results across INR architectures.
The high-resolution image is represented as a continuous coordinate-based
function. This formulation requires the optimizer to recover high-frequency
details from downsampled observations. The proposed optimizer achieves the
best or tied-best PSNR across the evaluated architectures. The results
support the proposed method as an effective update rule for INR-based
inverse reconstruction.

\vspace{-0.17in}
\paragraph{Neural Radiance Fields (NeRF).}
We evaluate the proposed optimizer on novel view synthesis with neural radiance fields~\cite{barron2021mip, mildenhall2021nerf, muller2022instant}.
NeRF represents a scene as a continuous volumetric function and renders images from queried camera rays.
This task requires optimization over geometry and appearance from multi-view observations. 
As shown in Table~\ref{tab:nerf_quan}, we conduct experiments on the Synthetic NeRF~\cite{mildenhall2021nerf} and LLFF datasets~\cite{mildenhall2019local}.
Each NeRF experiment uses the default hyperparameter configuration of its corresponding baseline implementation, and the proposed optimizer is matched to the same configuration as Adam. For validating progressive rank update, MoFaSGD~\cite{mahdavinia2025low} (fixed \& low-rank subspace update) and  Muon~\cite{jordan2024muon} (full-rank subspace update) are compared.

\begin{figure}[t]
\hfill
\begin{subfigure}{0.242\linewidth}
    \centering
    \DinoZoom{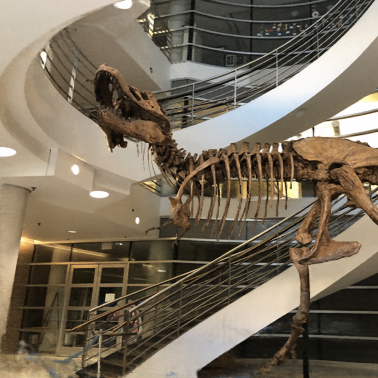}{\linewidth}{Muon}
\end{subfigure}
\hfill
\begin{subfigure}{0.242\linewidth}
    \centering
    \DinoZoom{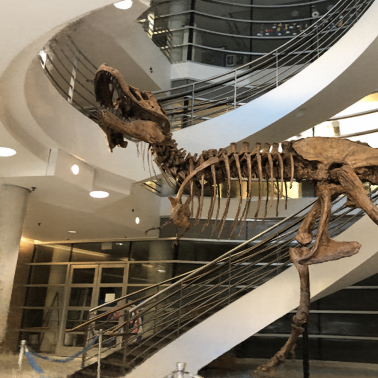}{\linewidth}{Ours}
\end{subfigure}
\hfill
\begin{subfigure}{0.242\linewidth}
    \centering
    \ShipZoom{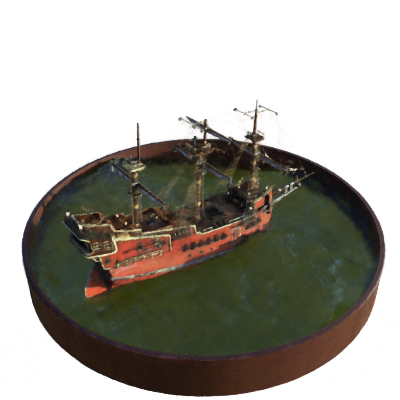}{\linewidth}{Muon}
\end{subfigure}
\hfill
\begin{subfigure}{0.242\linewidth}
    \centering
    \ShipZoom{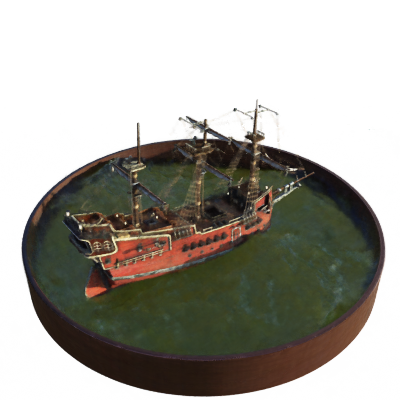}{\linewidth}{Ours}
\end{subfigure}
\hfill
\vspace{-0.05in}
\caption{\small \textbf{Qualitative comparison on NeRF~\cite{mildenhall2021nerf} scenes.} 
On both the real-world \textit{Trex} (LLFF) and the synthetic \textit{Ship} 
(Blender) scenes, our method produces sharper fine-grained details than Muon~\cite{jordan2024muon}, 
as highlighted in the zoom-in.}
\label{fig:nerf_qual}
\vspace{-0.24in}
\end{figure}

Figure~\ref{fig:nerf_qual} shows that our method recovers finer details. Table~\ref{tab:mipnerf_quan} reports reconstruction metrics on Mip-NeRF.
Table~\ref{tab:nerf_quan} reports scene-level results for vanilla NeRF on the LLFF and Synthetic NeRF datasets at 100k iterations; results at the original 200k iteration budget are in Appendix~\ref{sec:nerf_200k_results}.
Table~\ref{tab:ingp_quan} reports reconstruction metrics on Instant-NGP under the 1-second training budget setting reported in~\cite{muller2022instant}.
For Tables~\ref{tab:nerf_quan}--\ref{tab:hpo_quan}, bold and underlined values indicate the best and second-best results, respectively. 
Gray cells report average scores, and blue-shaded cells denote the proposed method.

On Mip-NeRF, the proposed optimizer improves reconstruction quality across synthetic scenes.
Table~\ref{tab:mipnerf_quan} shows higher average scores than AdamW on Synthetic NeRF scenes.
Instant-NGP results show consistent gains across all synthetic scenes.
These results indicate that the proposed optimizer improves reconstruction quality across volumetric scene representations.

\begin{table}[t]
\centering
\providecommand{\avgcell}[1]{\cellcolor{gray!12}#1}
\providecommand{\avgcellbold}[1]{\cellcolor{gray!12}\textbf{#1}}
\vspace{-10pt}
\centering
\footnotesize

\captionsetup{
    width=\linewidth,
    margin=0pt,
    justification=raggedright,
    singlelinecheck=false,
    format=plain
}

\captionof{table}{\small\textbf{Quantitative comparison on vanilla NeRF~\cite{mildenhall2021nerf}.} 
Scene-level PSNR and SSIM are reported on the LLFF and Synthetic NeRF datasets. Experiments use the original hyperparameter configuration with 100k iterations.}
\vspace{0.03in}
\label{tab:nerf_quan}

{%
\renewcommand{\arraystretch}{0.94}%
\setlength{\tabcolsep}{2.2pt}%
\setlength{\aboverulesep}{0.15pt}%
\setlength{\belowrulesep}{0.15pt}%
\setlength{\cmidrulesep}{0.15pt}%

\makebox[\linewidth][c]{%
\begin{minipage}{0.9\linewidth}
\centering

\resizebox{0.92\linewidth}{!}{%
\begin{tabular}{@{}cc*{8}{c}c@{}}
\toprule
\multirow{2}{*}{\textbf{Metric}} & \multirow{2}{*}{\textbf{Method}}
& \multicolumn{8}{c}{\textbf{LLFF}}
& \\
\cmidrule(lr){3-10}
& & Fern & Flower & Fortress & Horns & Leaves & Orchids & Room & T-Rex
& \avgcell{\textcolor{gray}{Avg.}} \\
\midrule

\multirow{4}{*}{PSNR $\uparrow$}
& Adam~\cite{kingma2014adam}
& 26.38 & 27.47 & 31.74 & 27.79 & 21.92 & 21.10 & 31.96 & 26.90
& \avgcell{26.91} \\
& MoFaSGD~\cite{mahdavinia2025low}
& 26.24 & 27.30 & 31.42 & 27.32 & 21.56 & 20.98 & 32.01 & 26.71
& \avgcell{26.69} \\
& Muon~\cite{jordan2024muon}
& \underline{26.84} & \underline{28.12} & \underline{32.21} & \underline{28.60} & \underline{22.16} & \underline{21.42} & \underline{33.17} & \underline{27.79}
& \avgcell{\underline{27.53}} \\
& \bestcell{Ours}
& \bestcell{\textbf{27.22}} & \bestcell{\textbf{28.68}} & \bestcell{\textbf{32.82}} & \bestcell{\textbf{29.02}}
& \bestcell{\textbf{22.54}} & \bestcell{\textbf{21.80}} & \bestcell{\textbf{33.72}} & \bestcell{\textbf{28.44}}
& \avgcellbold{\textbf{28.03}} \\

\midrule

\multirow{4}{*}{SSIM $\uparrow$}
& Adam~\cite{kingma2014adam}
& 0.8322 & 0.8566 & 0.9036 & 0.8702 & 0.7737 & 0.7135 & 0.9515 & 0.9000
& \avgcell{0.8501} \\
& MoFaSGD~\cite{mahdavinia2025low}
& 0.8200 & 0.8435 & 0.8946 & 0.8489 & 0.7454 & 0.6934 & 0.9481 & 0.8884
& \avgcell{0.8352} \\
& Muon~\cite{jordan2024muon}
& \underline{0.8507} & \underline{0.8777} & \underline{0.9184} & \underline{0.8934} & \underline{0.7880} & \underline{0.7318} & \underline{0.9604} & \underline{0.9175}
& \avgcell{\underline{0.8672}} \\
& \bestcell{Ours}
& \bestcell{\textbf{0.8584}} & \bestcell{\textbf{0.8880}} & \bestcell{\textbf{0.9270}} & \bestcell{\textbf{0.8984}}
& \bestcell{\textbf{0.8014}} & \bestcell{\textbf{0.7508}} & \bestcell{\textbf{0.9629}} & \bestcell{\textbf{0.9280}}
& \avgcellbold{\textbf{0.8768}} \\
\bottomrule
\end{tabular}%
}

\vspace{1pt}

\resizebox{0.92\linewidth}{!}{%
\begin{tabular}{@{}cc*{8}{c}c@{}}
\toprule
\multirow{2}{*}{\textbf{Metric}} & \multirow{2}{*}{\textbf{Method}}
& \multicolumn{8}{c}{\textbf{Synthetic}}
& \\
\cmidrule(lr){3-10}
& & Chair & Drums & Ficus & Hotdog & Lego & Materials & Mic & Ship
& \avgcell{\textcolor{gray}{Avg.}} \\
\midrule

\multirow{4}{*}{PSNR $\uparrow$}
& Adam~\cite{kingma2014adam}
& 32.57 & 25.08 & 28.02 & 35.40 & 30.23 & 28.32 & 32.04 & 28.47
& \avgcell{30.02} \\
& MoFaSGD~\cite{mahdavinia2025low}
& 31.57 & 24.76 & 27.45 & 34.77 & 29.90 & 28.51 & 31.02 & 28.17
& \avgcell{29.26} \\
& Muon~\cite{jordan2024muon}
& \underline{33.22} & \underline{25.22} & \underline{28.28} & \underline{36.23} & \underline{30.95} & \underline{29.19} & \underline{32.64} & \underline{28.87}
& \avgcell{\underline{30.58}} \\
& \bestcell{Ours}
& \bestcell{\textbf{34.28}} & \bestcell{\textbf{25.72}} & \bestcell{\textbf{29.30}} & \bestcell{\textbf{37.36}}
& \bestcell{\textbf{31.94}} & \bestcell{\textbf{29.95}} & \bestcell{\textbf{33.59}} & \bestcell{\textbf{29.57}}
& \avgcellbold{\textbf{31.46}} \\

\midrule

\multirow{4}{*}{SSIM $\uparrow$}
& Adam~\cite{kingma2014adam}
& 0.9702 & 0.9217 & 0.9566 & 0.9757 & 0.9536 & 0.9483 & 0.9742 & 0.8576
& \avgcell{0.9447} \\
& MoFaSGD~\cite{mahdavinia2025low}
& 0.9600 & 0.9165 & 0.9497 & 0.9719 & 0.9488 & 0.9501 & 0.9688 & 0.8569
& \avgcell{0.9374} \\
& Muon~\cite{jordan2024muon}
& \underline{0.9747} & \underline{0.9242} & \underline{0.9602} & \underline{0.9790} & \underline{0.9622} & \underline{0.9585} & \underline{0.9775} & \underline{0.8683}
& \avgcell{\underline{0.9506}} \\
& \bestcell{Ours}
& \bestcell{\textbf{0.9791}} & \bestcell{\textbf{0.9304}} & \bestcell{\textbf{0.9669}} & \bestcell{\textbf{0.9817}}
& \bestcell{\textbf{0.9681}} & \bestcell{\textbf{0.9628}} & \bestcell{\textbf{0.9800}} & \bestcell{\textbf{0.8773}}
& \avgcellbold{\textbf{0.9558}} \\
\bottomrule
\end{tabular}%
}
\end{minipage}%
}%
}

\vspace{6pt}
\providecommand{\avgcell}[1]{\cellcolor{gray!12}#1}
\providecommand{\avgcellbold}[1]{\cellcolor{gray!12}\textbf{#1}}

\vspace{-0.1in}
\centering
\small

\captionsetup{
    width=\linewidth,
    margin=0pt,
    justification=raggedright,
    singlelinecheck=false,
    format=plain,
    position=top,
    skip=1pt
}

\captionof{table}{\small\textbf{Quantitative comparison on Mip-NeRF~\cite{barron2021mip}.}
PSNR and SSIM on Synthetic NeRF scenes, trained for 200k iterations.}
\label{tab:mipnerf_quan}

{%
\renewcommand{\arraystretch}{0.99}%
\setlength{\tabcolsep}{2.2pt}%
\setlength{\aboverulesep}{0.15pt}%
\setlength{\belowrulesep}{0.15pt}%
\setlength{\cmidrulesep}{0.15pt}%

\resizebox{0.84\linewidth}{!}{%
\begin{tabular}{@{}cc*{8}{c}c@{}}
\toprule

\multirow{2}{*}{\textbf{Metric}}
& \multirow{2}{*}{\textbf{Method}}
& \multicolumn{8}{c}{\textbf{Synthetic}}
& \\
\cmidrule(lr){3-10}

& & Chair & Drums & Ficus & Hotdog & Lego & Materials & Mic & Ship
& \avgcell{\textcolor{gray}{Avg.}} \\

\midrule

\multirow{3}{*}{PSNR $\uparrow$}
& AdamW~\cite{loshchilov2017decoupled}
& \underline{35.24} & \underline{25.64} & \underline{29.50} & \underline{37.75} & \underline{33.07} & \underline{30.55} & \underline{34.35} & \underline{29.95}
& \avgcell{\underline{32.01}} \\
& Muon~\cite{jordan2024muon}
& 34.68 & 25.51 & 29.17 & 37.63 & 32.45 & 30.25 & 34.30 & 29.78
& \avgcell{31.72} \\
& \bestcell{Ours}
& \bestcell{\textbf{35.58}}
& \bestcell{\textbf{25.71}}
& \bestcell{\textbf{29.67}}
& \bestcell{\textbf{38.25}}
& \bestcell{\textbf{33.56}}
& \bestcell{\textbf{31.12}}
& \bestcell{\textbf{34.91}}
& \bestcell{\textbf{30.28}}
& \avgcellbold{32.39} \\

\midrule

\multirow{3}{*}{SSIM $\uparrow$}
& AdamW~\cite{loshchilov2017decoupled}
& \underline{0.9829} & \textbf{0.9292} & \underline{0.9675} & \underline{0.9831} & \underline{0.9743} & \underline{0.9653} & \underline{0.9814} & \underline{0.8784}
& \avgcell{\underline{0.9578}} \\
& Muon~\cite{jordan2024muon}
& 0.9799 & 0.9239 & 0.9658 & 0.9818 & 0.9693 & 0.9620 & 0.9807 & 0.8746
& \avgcell{0.9548} \\
& \bestcell{Ours}
& \bestcell{\textbf{0.9839}}
& \bestcell{\underline{0.9289}}
& \bestcell{\textbf{0.9698}}
& \bestcell{\textbf{0.9848}}
& \bestcell{\textbf{0.9770}}
& \bestcell{\textbf{0.9691}}
& \bestcell{\textbf{0.9838}}
& \bestcell{\textbf{0.8866}}
& \avgcellbold{0.9605} \\

\bottomrule
\end{tabular}%
}%
}
\vspace{-0.1in}

\vspace{6pt}
\providecommand{\avgcell}[1]{\cellcolor{gray!12}#1}
\providecommand{\avgcellbold}[1]{\cellcolor{gray!12}\textbf{#1}}
\vspace{-0.1pt}
\centering
\footnotesize
\captionsetup{
    width=\linewidth,
    margin=0pt,
    justification=raggedright,
    singlelinecheck=false,
    format=plain,
    position=top,
    skip=0pt
}

\captionof{table}{\small\textbf{Quantitative comparison on Instant-NGP~\cite{muller2022instant}.}
PSNR and SSIM are reported on Synthetic NeRF scenes.}
\label{tab:ingp_quan}

{%
\renewcommand{\arraystretch}{0.94}%
\setlength{\tabcolsep}{2.2pt}%
\setlength{\aboverulesep}{0.15pt}%
\setlength{\belowrulesep}{0.15pt}%
\setlength{\cmidrulesep}{0.15pt}%

\resizebox{0.84\linewidth}{!}{%
\begin{tabular}{@{}cc*{8}{c}c@{}}
\toprule

\multirow{2}{*}{\textbf{Metric}}
& \multirow{2}{*}{\textbf{Method}}
& \multicolumn{8}{c}{\textbf{Synthetic}}
& \\
\cmidrule(lr){3-10}

& & Chair & Drums & Ficus & Hotdog & Lego & Materials & Mic & Ship
& \avgcell{\textcolor{gray}{Avg.}} \\

\midrule

\multirow{3}{*}{PSNR $\uparrow$}
& Adam~\cite{kingma2014adam}
& \underline{29.41} & \underline{21.90} & \underline{26.95} & 32.19 & \underline{28.24} & 25.35 & \underline{29.69} & 24.27
& \avgcell{\underline{27.25}} \\
& Muon~\cite{jordan2024muon}
& 28.79 & 21.73 & 21.91 & \underline{32.37} & 27.02 & \underline{25.80} & 24.13 & \underline{25.04}
& \avgcell{25.84} \\
& \bestcell{Ours}
& \bestcell{\textbf{30.04}}
& \bestcell{\textbf{23.08}}
& \bestcell{\textbf{27.01}}
& \bestcell{\textbf{33.83}}
& \bestcell{\textbf{29.05}}
& \bestcell{\textbf{27.51}}
& \bestcell{\textbf{30.28}}
& \bestcell{\textbf{26.69}}
& \avgcellbold{28.44} \\
\midrule
\multirow{3}{*}{SSIM $\uparrow$}
& Adam~\cite{kingma2014adam}
& \underline{0.9373} & 0.8598 & \underline{0.9317} & \underline{0.9629} & \underline{0.9284} & 0.8861 & \underline{0.9642} & \underline{0.8311}
& \avgcell{\underline{0.9127}} \\
& Muon~\cite{jordan2024muon}
& 0.9270 & \underline{0.8813} & 0.8730 & 0.9555 & 0.9013 & \underline{0.9095} & 0.9472 & 0.8101
& \avgcell{0.9006} \\
& \bestcell{Ours}
& \bestcell{\textbf{0.9420}}
& \bestcell{\textbf{0.8980}}
& \bestcell{\textbf{0.9405}}
& \bestcell{\textbf{0.9646}}
& \bestcell{\textbf{0.9338}}
& \bestcell{\textbf{0.9286}}
& \bestcell{\textbf{0.9689}}
& \bestcell{\textbf{0.8412}}
& \avgcellbold{0.9272} \\
\bottomrule
\end{tabular}%
}%
\vspace{-0.2in}
}

\end{table}

\vspace{0.15in}
\subsection{Hyperparameter Optimization (HPO)}
\vspace{-0.1in}

In the NeRF comparison, all optimizers are evaluated under the same training protocol with fixed hyperparameters. Coordinate-based neural representations such as NeRF~\cite{mildenhall2021nerf} are sensitive to optimization settings~\cite{tancik2021learned}, and empirical optimizer comparisons depend on the tuning protocol~\cite{choi2019empirical}. A performance gap under a single fixed configuration is therefore partly attributable to hyperparameter mismatch rather than the optimizer design alone. To reduce this source of bias, we conduct an HPO comparison for the NeRF experiments under a unified search budget.

Each HPO trial trains NeRF for 100K iterations under a sampled optimizer configuration and uses the validation-set PSNR as the objective. Since each trial requires a full training run, we adopt the Tree-structured Parzen Estimator (TPE)~\cite{bergstra2011algorithms} implemented in Optuna~\cite{akiba2019optuna}, which models the densities of high- and low-performing trials and samples candidates likely to improve validation performance.
Each optimizer receives the same trial budget, and the configuration with the highest validation PSNR is evaluated once on the test set using PSNR and SSIM. For Adam, the search variables are the learning rate and learning-rate decay. For Muon and the proposed optimizer, the search variables are Adam-branch learning rate, learning-rate decay, and the Muon/projected-polar learning rate.

The resulting HPO comparisons are reported in Table~\ref{tab:hpo_quan}.
After tuning, the proposed optimizer maintains the best average reconstruction quality, suggesting that the gain is driven by the optimizer design rather than unfavorable baseline hyperparameters.
Search spaces are provided in Appendix~\ref{sec:hpo_details}.

\ifdefined\hyperLeftBlockW\else\newlength{\hyperLeftBlockW}\fi
\ifdefined\hyperRightBlockW\else\newlength{\hyperRightBlockW}\fi

\begin{table}[t]
\vspace{-0.05in}
\centering
\small
\setlength{\hyperLeftBlockW}{0.63\linewidth}
\setlength{\hyperRightBlockW}{0.36\linewidth}

\captionsetup{
    width=\linewidth,
    margin=0pt,
    justification=raggedright,
    singlelinecheck=false,
    format=plain
}
    \captionof{table}{\small\textbf{NeRF hyperparameter tuning performance at 100K iterations.}
    Comparison of Adam~\cite{kingma2014adam}, Muon~\cite{jordan2024muon}, and Ours under fixed-hyperparameter and TPE-tuned~\cite{bergstra2011algorithms} settings on Synthetic NeRF scenes.}
    \vspace{2pt}
    \label{tab:hpo_quan}
    {%
    \renewcommand{\arraystretch}{1.12}%
    \setlength{\tabcolsep}{2.2pt}%
    \setlength{\aboverulesep}{0.15pt}%
    \setlength{\belowrulesep}{0.15pt}%
    \setlength{\cmidrulesep}{0.15pt}%

    \resizebox{0.97\linewidth}{!}{%
    \begin{tabular}{@{}cc*{4}{ccc}@{}}
    \toprule
    \multicolumn{2}{c}{\textbf{Evaluation}}
        & \multicolumn{12}{c}{\textbf{Synthetic}} \\
    \cmidrule(lr){1-2}
    \cmidrule(lr){3-14}

    \multirow[c]{2}{*}{\textbf{Metric}} 
    & \multirow[c]{2}{*}{\textbf{HPO setting}}
        & \multicolumn{3}{c}{\textbf{Chair}}
        & \multicolumn{3}{c}{\textbf{Drums}}
        & \multicolumn{3}{c}{\textbf{Ficus}}
        & \multicolumn{3}{c}{\textbf{Hotdog}} \\
    \cmidrule(lr){3-5}
    \cmidrule(lr){6-8}
    \cmidrule(lr){9-11}
    \cmidrule(lr){12-14}

    & 
        & \textbf{Adam} & \textbf{Muon} & \bestcell{\textbf{Ours}}
        & \textbf{Adam} & \textbf{Muon} & \bestcell{\textbf{Ours}}
        & \textbf{Adam} & \textbf{Muon} & \bestcell{\textbf{Ours}}
        & \textbf{Adam} & \textbf{Muon} & \bestcell{\textbf{Ours}} \\
    \midrule

    \multirow[c]{2}{*}{PSNR $\uparrow$}
        & Fixed HP & 32.57 & \underline{33.22} & \bestcell{\textbf{34.28}} & 25.08 & \underline{25.22} & \bestcell{\textbf{25.72}} & 28.02 & \underline{28.28} & \bestcell{\textbf{29.30}} & 35.40 & \underline{36.23} & \bestcell{\textbf{37.36}} \\
        & TPE & 32.93 & \underline{34.19} & \bestcell{\textbf{34.37}} & 25.20 & \textbf{25.73} & \bestcell{\underline{25.61}} & 28.33 & \textbf{29.30} & \bestcell{\textbf{29.30}} & 37.35 & \underline{37.40} & \bestcell{\textbf{37.48}} \\
    \midrule

    \multirow[c]{2}{*}{SSIM $\uparrow$}
        & Fixed HP & 0.9702 & \underline{0.9747} & \bestcell{\textbf{0.9791}} & 0.9217 & \underline{0.9242} & \bestcell{\textbf{0.9304}} & 0.9566 & \underline{0.9602} & \bestcell{\textbf{0.9669}} & 0.9757 & \underline{0.9790} & \bestcell{\textbf{0.9817}} \\
        & TPE & 0.9702 & \underline{0.9773} & \bestcell{\textbf{0.9793}} & 0.9259 & \textbf{0.9317} & \bestcell{\underline{0.9293}} & 0.9568 & \underline{0.9656} & \bestcell{\textbf{0.9666}} & 0.9775 & \textbf{0.9821} & \bestcell{\textbf{0.9821}} \\

    \midrule[0.7pt]

    \multirow[c]{2}{*}{\textbf{Metric}} 
    & \multirow[c]{2}{*}{\textbf{HPO setting}}
        & \multicolumn{3}{c}{\textbf{Lego}}
        & \multicolumn{3}{c}{\textbf{Materials}}
        & \multicolumn{3}{c}{\textbf{Mic}}
        & \multicolumn{3}{c}{\textbf{Ship}} \\
    \cmidrule(lr){3-5}
    \cmidrule(lr){6-8}
    \cmidrule(lr){9-11}
    \cmidrule(lr){12-14}

    & 
        & \textbf{Adam} & \textbf{Muon} & \bestcell{\textbf{Ours}}
        & \textbf{Adam} & \textbf{Muon} & \bestcell{\textbf{Ours}}
        & \textbf{Adam} & \textbf{Muon} & \bestcell{\textbf{Ours}}
        & \textbf{Adam} & \textbf{Muon} & \bestcell{\textbf{Ours}} \\
    \midrule

    \multirow[c]{2}{*}{PSNR $\uparrow$}
        & Fixed HP & 30.23 & \underline{30.95} & \bestcell{\textbf{31.94}} & 28.32 & \underline{29.19} & \bestcell{\textbf{29.95}} & 32.04 & \underline{32.64} & \bestcell{\textbf{33.59}}  & 28.47 & \underline{28.87} & \bestcell{\textbf{29.57}} \\
        & TPE  & 30.61 & \underline{31.66} & \bestcell{\textbf{32.03}} & 28.37 & \underline{29.55} & \bestcell{\textbf{29.95}} & 32.62 & \underline{33.64} & \bestcell{\textbf{33.71}} & 28.51 & \underline{29.48} & \bestcell{\textbf{29.64}} \\

    \midrule

    \multirow[c]{2}{*}{SSIM $\uparrow$}
        & Fixed HP & 0.9536 & \underline{0.9622} & \bestcell{\textbf{0.9681}} & 0.9483 & \underline{0.9585} & \bestcell{\textbf{0.9628}} & 0.9742 & \underline{0.9775} & \bestcell{\textbf{0.9800}} & 0.8576 & \underline{0.8683} & \bestcell{\textbf{0.8773}} \\
        & TPE  & 0.9552 & \underline{0.9651} & \bestcell{\textbf{0.9688}} & 0.9485 & \underline{0.9597} & \bestcell{\textbf{0.9627}} & 0.9750 & \underline{0.9802} & \bestcell{\textbf{0.9804}} & 0.8618 & \textbf{0.8813} & \bestcell{\underline{0.8774}} \\

\bottomrule
\end{tabular}%
}%
}
\vspace{-0.22in}
\end{table}

\vspace{-0.17in}
\section{Analysis}
\vspace{-0.15in}
\label{sec:analysis}

In this section, we analyze the rank dynamics of the momentum matrix
\(M_t\) and the parameter update \(\Delta W_t = W_{t+1}-W_t\), and then
examine how different components of the polar factor of \(M_t\) affect
frequency-band residual reduction during INR training. In Section~\ref{sec:rank_analysis},
we first measure the normalized entropy effective rank of \(M_t\) and
\(\Delta W_t\), and compare rank-schedule configurations to examine whether increasing the rank over training better matches this observed behavior.
In Section~\ref{sec:tail_analysis}, we decompose the full polar factor \(\bar M_t\) into
head and tail components and evaluate their effects on residual energy
in low-, mid-, and high-frequency bands during image overfitting.

\vspace{-0.07in}
\subsection{Effective Rank and Progressively Increasing Rank}
\vspace{-0.02in}
\label{sec:rank_analysis}

\begin{wrapfigure}[25]{r}{0.45\linewidth}
    \vspace{-0.23in}
    \centering
    \includegraphics[width=1\linewidth]{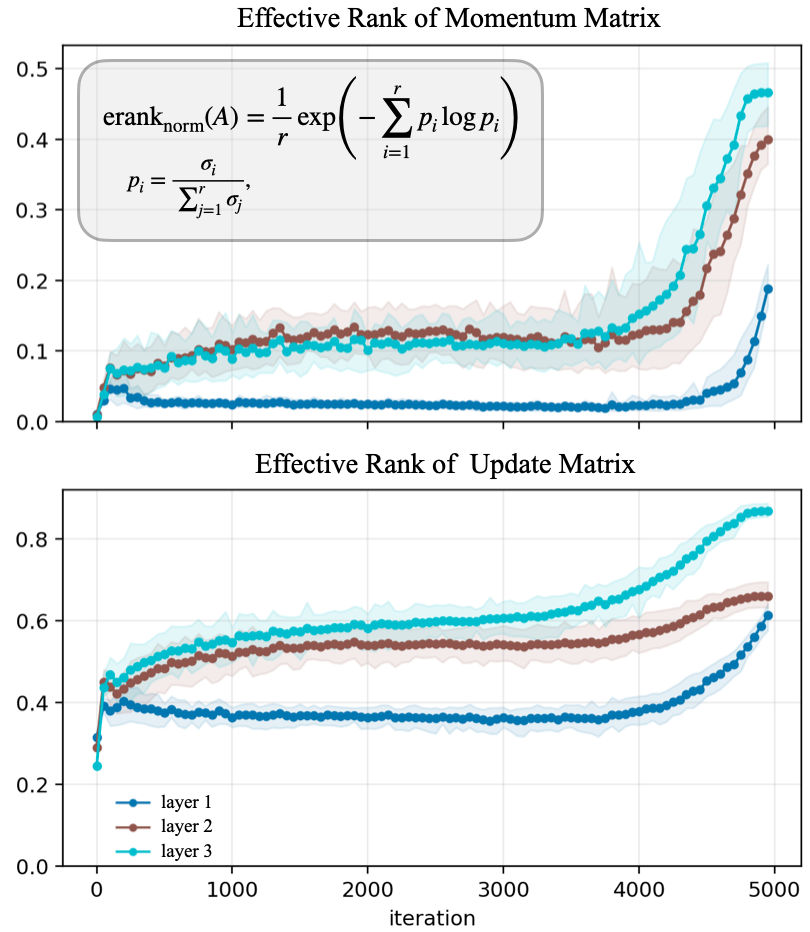}
    \vspace{-0.25in}
    \caption{\small\textbf{Effective-rank dynamics during INR training.}
    Normalized entropy effective rank of the momentum matrix \(M_t\) (top) and the parameter update \(\Delta W_t = W_{t+1}-W_t\) (bottom). 
    The normalized effective rank increases in later iterations, indicating that more singular components become non-negligible as training proceeds.}
    \label{fig:rank_plot}
\end{wrapfigure}
To examine whether the effective rank~\cite{roy2007effective} of the momentum matrices increases during training, we monitor their normalized entropy effective rank. This quantity is defined in Figure~\ref{fig:rank_plot}. A low value indicates that a small number of leading singular values account for a large fraction of the sum of singular values. Figure~\ref{fig:rank_plot} presents the evolution of the normalized entropy effective rank of the momentum matrices \(M_t\) and the parameter updates \(\Delta W_t\). For most of training, both effective-rank measures remain low, then increase rapidly in the later iterations. This behavior suggests that, in the early stage of training, a low-rank approximation is sufficient for the momentum matrices and parameter updates. In later iterations, more singular values become non-negligible, supporting the use of a higher rank.

\begin{figure}[t]
\centering
\small
\setlength{\tabcolsep}{0pt}
\renewcommand{\arraystretch}{1}
\begin{tabular}{@{}c@{\hspace{1pt}}c@{\hspace{1pt}}c@{\hspace{1pt}}c@{\hspace{1pt}}c@{}}
    \begin{tabular}{@{}c@{}}
        \AbGTZoom{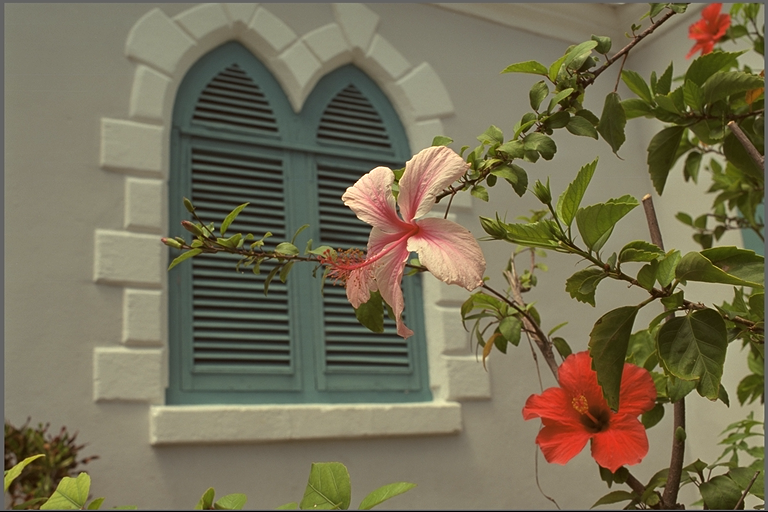}{0.20\linewidth}{}
        \\
        {\small Ground Truth}
    \end{tabular}
    &
    \begin{tabular}{@{}c@{}}
        \PSNRZoomImage{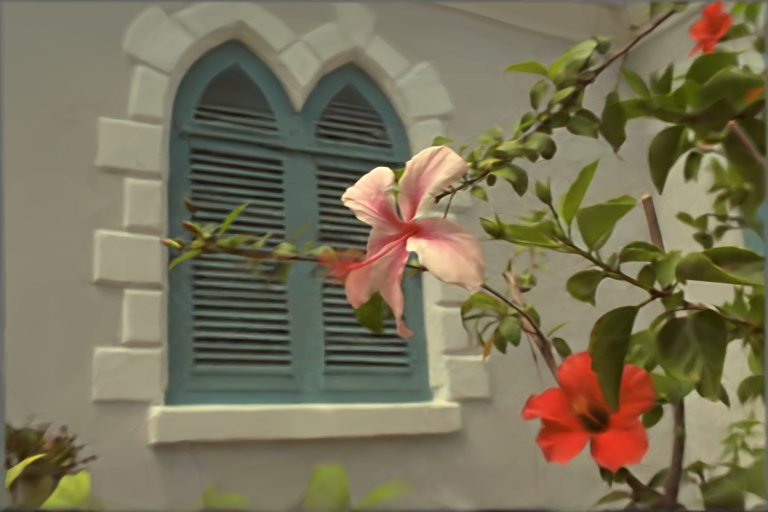}{0.20\linewidth}{27.12}{-0.08}{0.53}{1.195}{-1.32}
        \\
        {\small Increasing rank}
    \end{tabular}
    &
    \begin{tabular}{@{}c@{}}
        \PSNRZoomImage{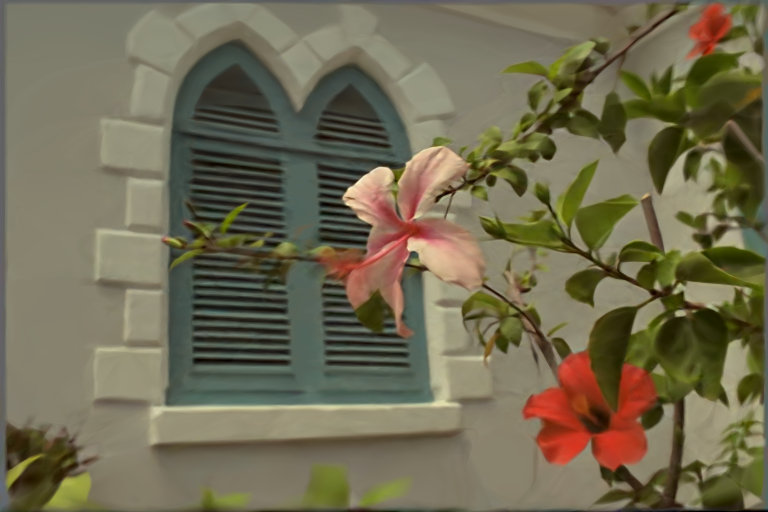}{0.20\linewidth}{26.58}{-0.08}{0.53}{1.195}{-1.32}
        \\
        {\small Fixed rank}
    \end{tabular}
    &
    \begin{tabular}{@{}c@{}}
        \PSNRZoomImage{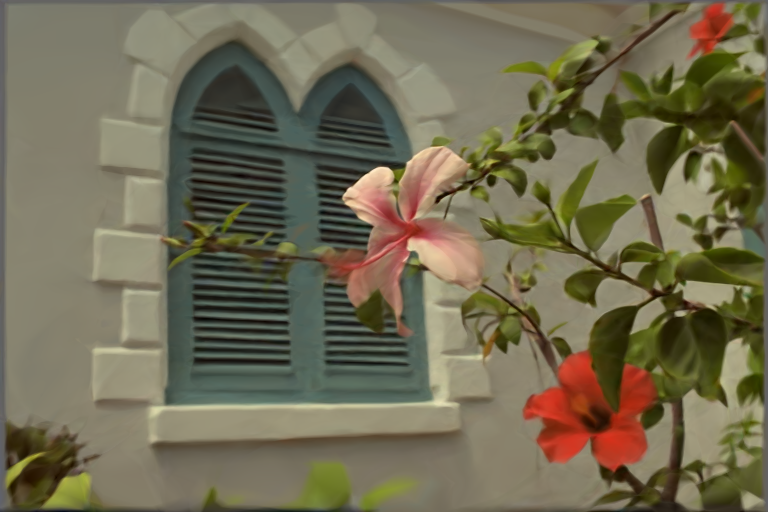}{0.20\linewidth}{26.47}{-0.08}{0.53}{1.195}{-1.32}
        \\
        {\small Full rank}
    \end{tabular}
    &
    \begin{tabular}{@{}c@{}}
        \PSNRZoomImage{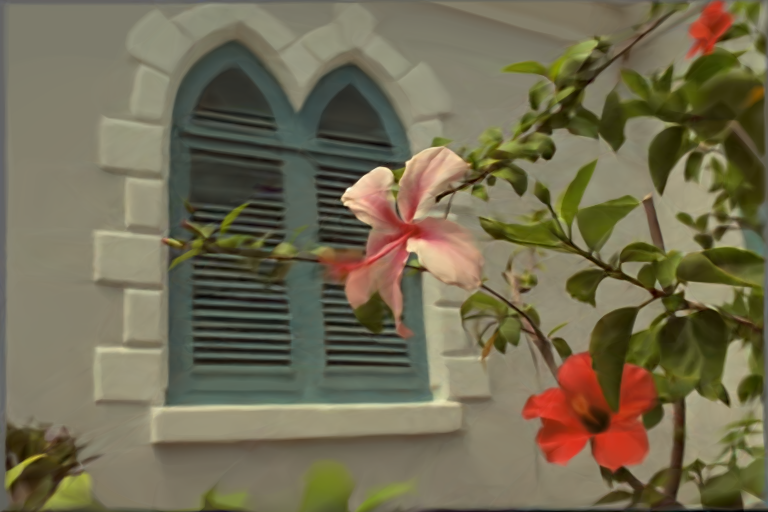}{0.20\linewidth}{26.06}{-0.08}{0.53}{1.195}{-1.32}
        \\
        {\small Decreasing rank}
    \end{tabular}
\end{tabular}
\captionsetup{
    width=0.95\linewidth,
    margin=0pt,
    justification=raggedright,
    singlelinecheck=true,
    skip=2pt
}
\captionof{figure}{\small \textbf{Rank-configuration comparison.}
Qualitative comparison of increasing-rank, fixed-rank, decreasing-rank and full-rank schedules on image overfitting. Further details are provided in Appendix~\ref{app:schedule}.}
\label{fig:curve_gt_comparison}
\vspace{-0.25in}
\end{figure}

Figure~\ref{fig:curve_gt_comparison} compares four configurations of the rank schedule in the image overfitting setting: increasing rank, full rank, fixed rank, and decreasing rank. Table~\ref{tab:fig6_evidence} reports the detailed quantitative results. The increasing-rank schedule achieves the best reconstruction quality, whereas schedules with a static or decreasing rank yield lower performance. This result supports the proposed increasing rank schedule, which increases rank as optimization proceeds instead of keeping it fixed throughout training. 

\vspace{-0.02in}
\subsection{Frequency-Band Analysis on Head and Tail matrices}
\label{sec:tail_analysis}
\pgfplotsset{compat=1.18}

\vspace{-0.05in}

Let \(M_t=U_t\Sigma_tV_t^\top\) be the singular value decomposition of the momentum matrix, and let 
\(\bar{M}_t=U_tV_t^\top\) denote its full polar factor. 
We take the left and right singular vectors associated with the \(k\) largest singular values, collect them as \(U_{t,k}\) and \(V_{t,k}\), and define
\(\bar{M}_{t,k}=U_{t,k}V_{t,k}^{\top}\). 
\begin{wrapfigure}{r}{0.42\textwidth}
\vspace{-10pt}
\centering
\small
\captionof{table}{\small\textbf{Rank schedule ablation.}
Mean $\pm$ std over the 24 images of Kodak24 under the same
learning-rate scheduler. Best results are in bold. PSNR at intermediate checkpoints is reported in Appendix~\ref{app:schedule}.}
\label{tab:fig6_evidence}
\vspace{-2pt}

\renewcommand{\arraystretch}{0.985}%
\setlength{\tabcolsep}{2pt}%
\setlength{\aboverulesep}{0.15ex}%
\setlength{\belowrulesep}{0.15ex}%
\newcommand{\mstd}[2]{#1\,$\pm$\,#2}%
\newcommand{\mstdbf}[2]{\textbf{#1}\,{\boldmath$\pm$}\,\textbf{#2}}%
\begin{tabular}{@{}lcc@{}}
\toprule[0.8pt]
~\textbf{Method} & \textbf{PSNR}\,$\uparrow$ & \textbf{SSIM}\,$\uparrow$ \\
\midrule
\scriptsize Decreasing rank         & \mstd{25.53}{6.45}   & \mstd{0.7014}{0.1136}   \\
\scriptsize Full rank    & \mstd{25.22}{7.08}   & \mstd{0.7018}{0.1444}   \\
\scriptsize Fixed rank          & \mstd{25.25}{5.13}   & \mstd{0.7095}{0.1234}   \\
\scriptsize \textbf{Increasing rank} & \mstdbf{27.33}{5.37} & \mstdbf{0.7363}{0.1210} \\
\bottomrule[0.8pt]
\end{tabular}
\vspace{-3pt}
\end{wrapfigure}

We refer to \(\bar{M}_{t,k}\) as the rank-\(k\) head component. 
The corresponding tail component is defined as the remaining part of the polar factor, \(\bar{M}_t-\bar{M}_{t,k}\).
We measure the effect of the head and tail components of \(\bar{M}_t\) on residual energy in low-, mid-, and high-frequency bands during image overfitting. For a frequency band \(B\), we define the mean residual power as
\(E_t(B)=\frac{1}{|B|C}\sum_{\omega\in B}\sum_{c=1}^{C}
|\mathcal{F}(R_t^{(c)})(\omega)|^2\), where \(R_t\) is the residual image,
\(C\) is the number of color channels, and \(\mathcal{F}\) denotes the
two-dimensional Fourier transform. For visualization, we report the
relative residual energy \(\widetilde{E}_t(B)=E_t(B)/E_0(B)\).
\vspace{-2pt}

\begin{wrapfigure}[14]{r}{0.59\textwidth}
\vspace{-0.2in}
\setlength{\abovecaptionskip}{2pt}
\setlength{\belowcaptionskip}{0pt}
\centering
\includegraphics[width=\linewidth]{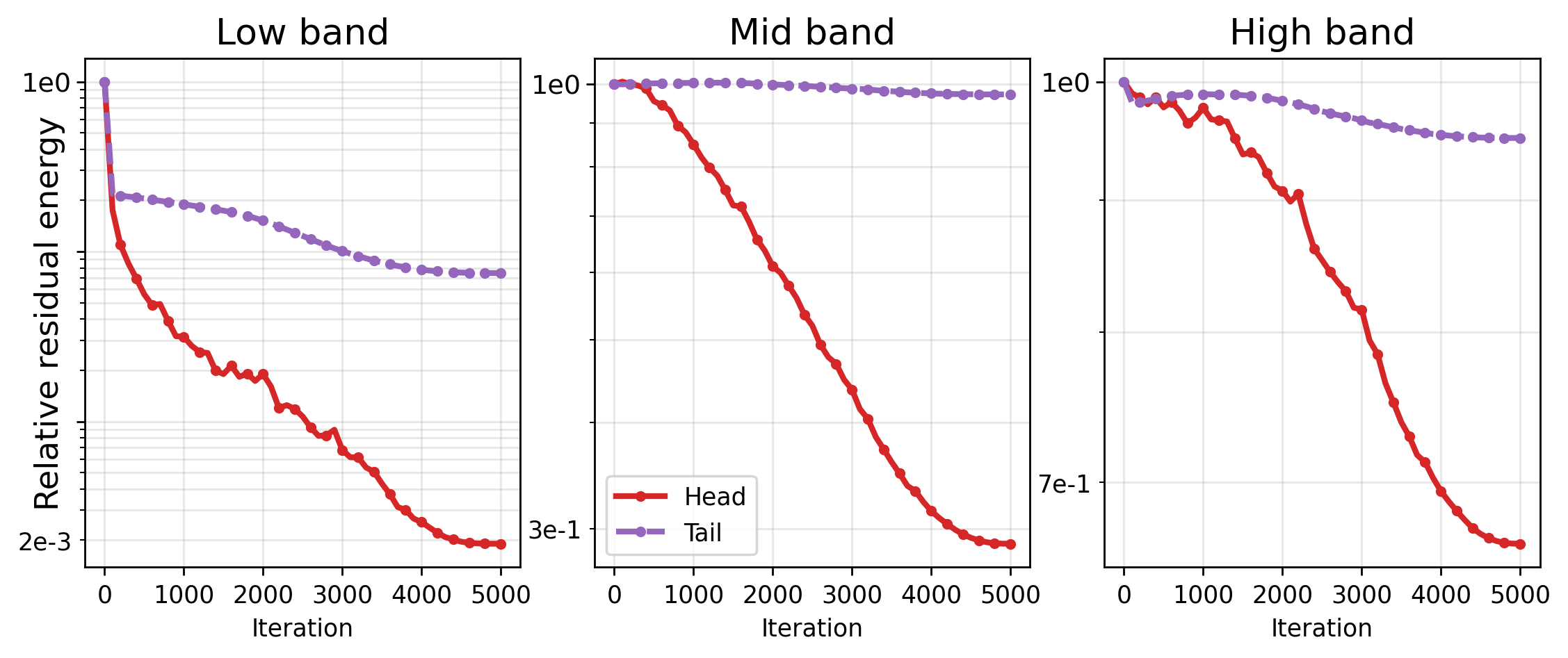}
\vspace{-0.18in}
\caption{\small\textbf{Frequency-band residual energy analysis of the head and tail terms.}
We compare the head and tail terms with \(k=128\) during 5000 iterations of image overfitting on Kodak24. The plots show relative residual energy in the low- (left), mid- (center), and high-frequency (right) bands.}
\label{fig:tail}
\vspace{-0.10in}
\end{wrapfigure}

\paragraph{Frequency-band trajectories.}
Figure~\ref{fig:tail} reports residual energy in three frequency bands under the head-only and tail-only settings for \(k=128\).  The head term reduces
low-frequency residual energy by orders of magnitude and continues to reduce
mid- and high-frequency residuals. In contrast, the tail term decreases slowly
in the low-frequency band and stays near its initial level in the other bands.
These results suggest that the update components responsible for fine-detail recovery are concentrated in the head term of \(\bar{M}_t\), rather than in the tail. Thus, the tail term alone does not account for high-frequency reconstruction.

\vspace{-3pt}
\begin{wrapfigure}[21]{r}{0.35\textwidth}
\vspace{-0.07in}
  \centering
  \begin{tabular}{c}
    \includegraphics[width=1\linewidth]{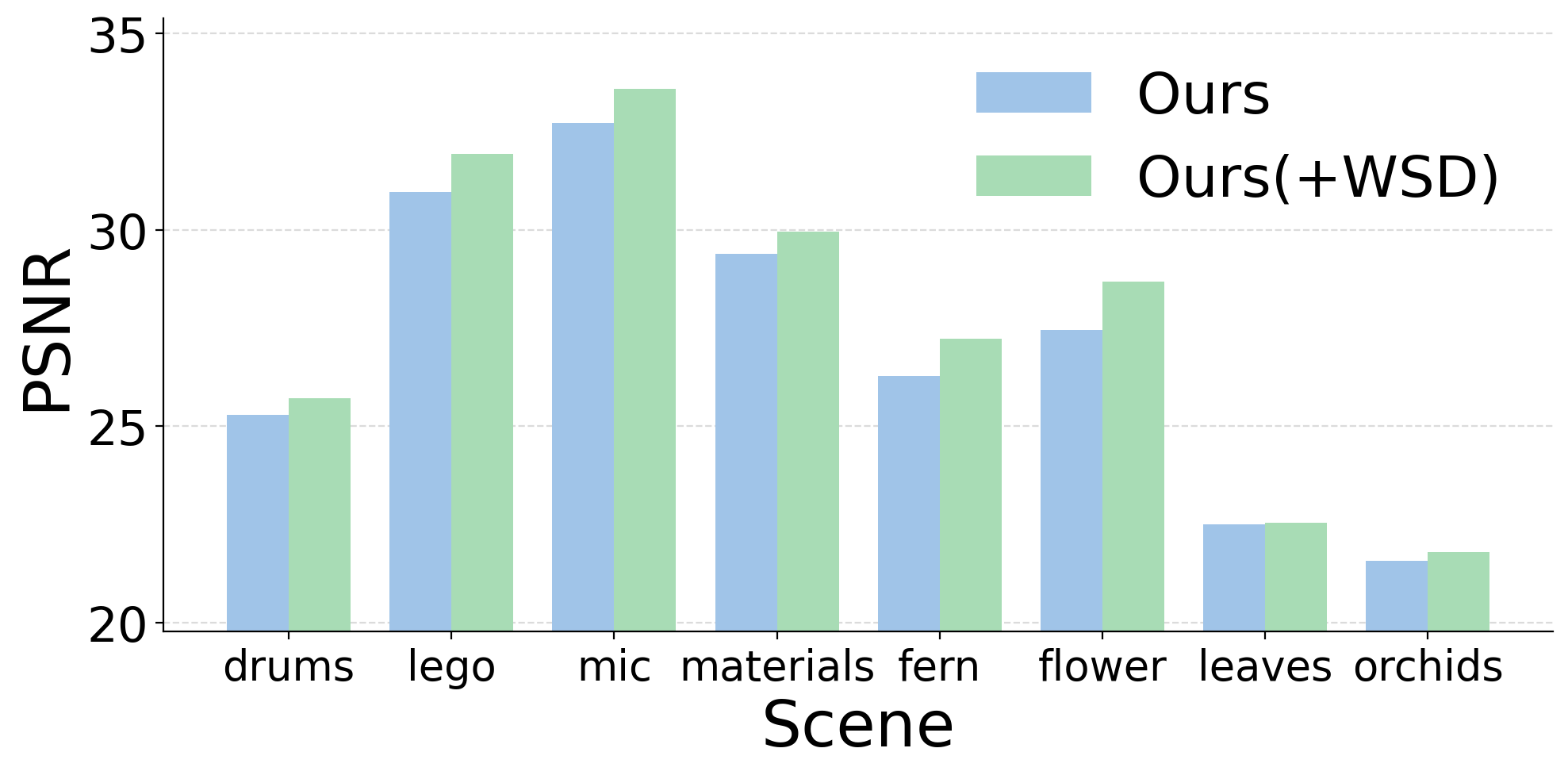} \\
    \includegraphics[width=1\linewidth]{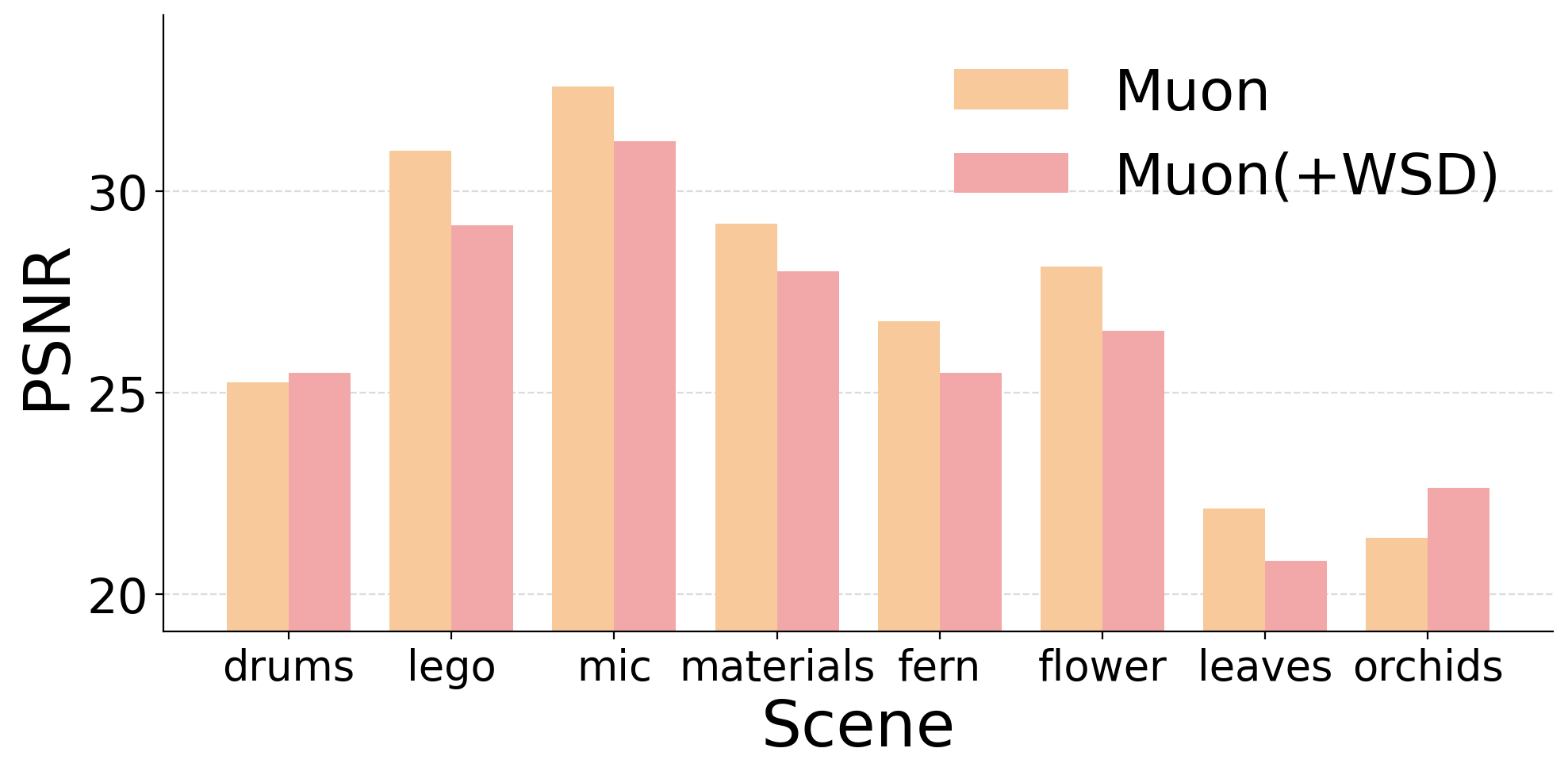} \\
  \end{tabular}
  \vspace{-0.1in}
  \caption{\small \textbf{Effect of the rank-aware WSD schedule.} PSNR across scenes with and without the WSD schedule, applied to the proposed optimizer (top) and the baseline (bottom), evaluated on  synthetic NeRF scenes.}
  \label{fig:wsd_ab}
  \vspace{-0.3in}

\end{wrapfigure}

\paragraph{Ablation study.}

We ablate the rank-aware WSD scheduler introduced in Section~\ref{subsec:rawsd}.  For each optimizer, we compare training with and without the WSD scheduler under the same training budget. Figure~\ref{fig:wsd_ab} shows that the WSD scheduler consistently improves the proposed optimizer across the learning-rate range compared with its non-WSD counterpart. This behavior indicates that WSD is effective when coupled with the proposed optimizer: by maintaining the peak step size during the rank-growth phase, the scheduler matches the learning-rate schedule to the progressive expansion of the retained leading left and right subspaces. This coupling is a distinguishing characteristic relative to the fixed-rank baseline.

\paragraph{Rank-growth schedule.}
We compare cosine, logistic, and piecewise rank-growth schedules under the same learning-rate scheduler and rank bounds in Table~\ref{tab:growth_shape}. The three schedules yield similar PSNR and SSIM. Together with Table~\ref{tab:fig6_evidence}, this result suggests that the increase in rank contributes more to reconstruction quality than the shape of the growth curve. We adopt the cosine schedule since it requires no additional shape hyperparameter.

\vspace{-3pt}

\newpage
\section{Limitations}
\vspace{-0.13in}
\paragraph{Training time.}
Similar to Muon~\cite{jordan2024muon}, our method incurs additional computation relative to Adam due to the polar decomposition step. Table~\ref{tab:nerf_training_time} reports the time per iteration and the optimizer-step time on the Lego scene after 50 warm-up iterations. In our preliminary experiments, periodically reusing the projection basis substantially reduced optimizer-step time with no noticeable loss of quality, suggesting a promising direction for further reducing this overhead.
\vspace{-0.13in}
\paragraph{Gain under tuning.}
In Table~\ref{tab:hpo_quan}, our method reaches near its tuned performance under the default configuration, and TPE tuning yields an additional gain of 0.05\,dB in average PSNR on Synthetic NeRF. This gain is narrow compared with those of the other optimizers.
\vspace{-0.13in}
\paragraph{Scope of evaluation.}
Our evaluation covers MLP-based coordinate INRs trained from scratch, and other recent INR architectures remain unexplored. Under a warm start from an initial encoding stage, the coarse structure is already fitted, and the benefit of the rank schedule is expected to decrease relative to its overhead.
\begin{table}[t]
\centering
\captionsetup{font=small, justification=raggedright, singlelinecheck=false}
\begin{minipage}[t]{0.47\linewidth}
\centering
\caption{\textbf{Rank-growth schedule.} Mean $\pm$ std over the 24 images of Kodak24 at 3,000 iterations. Experiments use a single NVIDIA RTX PRO 6000 GPU.}
\label{tab:growth_shape}
\small
\begin{tabular*}{\linewidth}{@{\extracolsep{\fill}}ccc@{}}
\toprule
~~~Schedule & PSNR\,$\uparrow$ & SSIM\,$\uparrow$~~ \\
\midrule
~~~Cosine    & 27.33\,{\scriptsize$\pm$\,5.38} & 0.7363\,{\scriptsize$\pm$\,0.1210}~~ \\
~~~Logistic  & 27.32\,{\scriptsize$\pm$\,4.15} & 0.7355\,{\scriptsize$\pm$\,0.1492}~~ \\
~~~Piecewise & 27.32\,{\scriptsize$\pm$\,4.18} & 0.7358\,{\scriptsize$\pm$\,0.1402}~~ \\
\bottomrule
\end{tabular*}
\end{minipage}%
\hfill
\begin{minipage}[t]{0.49\linewidth}
\centering
\caption{\textbf{Training time comparison on the Synthetic NeRF Lego scene.} Iter and Step denote the time per iteration and the \texttt{optimizer.step()} time, averaged over 10K iterations.}
\label{tab:nerf_training_time}
\small
\begin{tabular*}{\linewidth}{@{\extracolsep{\fill}}cccc@{}}
\toprule
~~~Method & Params & Iter (ms) & Step (ms)~~ \\
\midrule
~~~Adam & \cellcolor{gray!10} & \phantom{0}96.63 & \phantom{0}5.23~~ \\
~~~Muon & \cellcolor{gray!10} & 107.22 & 19.75~~ \\
~~~Ours & \cellcolor{gray!10}\multirow{-3}{*}{1.19M} & 112.70 & 25.45~~ \\
\bottomrule
\end{tabular*}
\end{minipage}
\vspace{-0.08in}
\end{table}

\vspace{-0.05in}
\section{Conclusion}
\label{sec:conclusion}
\vspace{-0.13in}
In this paper, we proposed Progressive Polar Decomposition, a robust optimizer designed for INR training. 
The proposed method projects matrix-valued momentum onto a low-rank orthonormal basis, applies polar normalization in the compressed coordinates, and reconstructs the update in the original parameter space, with an energy-calibrated cosine schedule that progressively expands the subspace rank during training. 
This design aligns the optimizer state with the coarse-to-fine learning behavior of INRs. 
We evaluated the proposed method on image fitting, single-image super-resolution, and neural radiance field optimization, where it consistently improved reconstruction quality across the evaluated INR architectures, with PSNR gains of up to 6\,dB on image fitting. 
For NeRF, we further conducted a unified HPO comparison to reduce the influence of optimizer-specific hyperparameter choices, and the proposed optimizer remained strongest under the tuned setting.

\begin{ack}
This work was partly supported by the National Research
Foundation of Korea(NRF) grant funded by the Korea government(MSIT) (RS-2024-00335741), 
Institute of Information \& communications Technology Planning \& Evaluation(IITP) grant funded by the Korea government(MSIT) (RS2025-25442405 Development of a Self-Learning World
Model-Based AGI System for Hyperspectral Imaging),
This work was supported in part by the National Research Foundation of Korea (NRF) grant funded by the Korea government (MSIT) (No. RS-2026-25499511), the Institute of Information \& Communications Technology Planning \& Evaluation (IITP) grant funded by the Korea government (MSIT) (No. RS-2025-02263080, Development of PIM Memory Technology for a High-Speed Interface-Based Programmable Computing Architecture).
\end{ack}

{
    \newpage
    \small
    \bibliographystyle{ieeenat_fullname}
    \bibliography{refs}

@String(CVPR= {IEEE Conf. Comput. Vis. Pattern Recog.})

@String(TOG= {ACM Trans. Graph.})

@String(CVPR  = {CVPR})

@String(TOG   = {ACM TOG})

@article{halko2011finding,
  title={Finding structure with randomness: Probabilistic algorithms for constructing approximate matrix decompositions},
  author={Halko, Nathan and Martinsson, Per-Gunnar and Tropp, Joel A},
  journal={SIAM review},
  volume={53},
  number={2},
  pages={217--288},
  year={2011},
  publisher={SIAM}
}

@inproceedings{shi2024improved,
  title={Improved implicit neural representation with fourier reparameterized training},
  author={Shi, Kexuan and Zhou, Xingyu and Gu, Shuhang},
  booktitle={Proceedings of the IEEE/CVF Conference on Computer Vision and Pattern Recognition},
  pages={25985--25994},
  year={2024}
}

@article{shi2024inductive,
  title={Inductive gradient adjustment for spectral bias in implicit neural representations},
  author={Shi, Kexuan and Chen, Hai and Zhang, Leheng and Gu, Shuhang},
  journal={arXiv preprint arXiv:2410.13271},
  year={2024}
}

@article{mcginnis2025optimizing,
  title={Optimizing Rank for High-Fidelity Implicit Neural Representations},
  author={McGinnis, Julian and H{\"o}lzl, Florian A and Shit, Suprosanna and Bieder, Florentin and Friedrich, Paul and M{\"u}hlau, Mark and Menze, Bj{\"o}rn and Rueckert, Daniel and Wiestler, Benedikt},
  journal={arXiv preprint arXiv:2512.14366},
  year={2025}
}

@article{zheng2026inlet,
  title={The Inlet Rank Collapse in Implicit Neural Representations: Diagnosis and Unified Remedy},
  author={Zheng, Jianqiao and Saratchandran, Hemanth and Lucey, Simon},
  journal={arXiv preprint arXiv:2602.01526},
  year={2026}
}

@inproceedings{rahaman2019spectral,
  title={On the spectral bias of neural networks},
  author={Rahaman, Nasim and Baratin, Aristide and Arpit, Devansh and Draxler, Felix and Lin, Min and Hamprecht, Fred and Bengio, Yoshua and Courville, Aaron},
  booktitle={International conference on machine learning},
  pages={5301--5310},
  year={2019},
  organization={PMLR}
}

@article{tancik2020fourier,
  title={Fourier features let networks learn high frequency functions in low dimensional domains},
  author={Tancik, Matthew and Srinivasan, Pratul and Mildenhall, Ben and Fridovich-Keil, Sara and Raghavan, Nithin and Singhal, Utkarsh and Ramamoorthi, Ravi and Barron, Jonathan and Ng, Ren},
  journal={Advances in neural information processing systems},
  volume={33},
  pages={7537--7547},
  year={2020}
}

@article{sitzmann2020implicit,
  title={Implicit neural representations with periodic activation functions},
  author={Sitzmann, Vincent and Martel, Julien and Bergman, Alexander and Lindell, David and Wetzstein, Gordon},
  journal={Advances in neural information processing systems},
  volume={33},
  pages={7462--7473},
  year={2020}
}

@inproceedings{liu2024finer,
  title={Finer: Flexible spectral-bias tuning in implicit neural representation by variable-periodic activation functions},
  author={Liu, Zhen and Zhu, Hao and Zhang, Qi and Fu, Jingde and Deng, Weibing and Ma, Zhan and Guo, Yanwen and Cao, Xun},
  booktitle={Proceedings of the IEEE/CVF Conference on Computer Vision and Pattern Recognition},
  pages={2713--2722},
  year={2024}
}

@misc{saragadam2023wirewaveletimplicitneural,
      title={WIRE: Wavelet Implicit Neural Representations}, 
      author={Vishwanath Saragadam and Daniel LeJeune and Jasper Tan and Guha Balakrishnan and Ashok Veeraraghavan and Richard G. Baraniuk},
      year={2023},
      eprint={2301.05187},
      archivePrefix={arXiv},
      primaryClass={cs.CV},
      url={https://arxiv.org/abs/2301.05187}, 
}

@article{hertz2021sape,
  title={Sape: Spatially-adaptive progressive encoding for neural optimization},
  author={Hertz, Amir and Perel, Or and Giryes, Raja and Sorkine-Hornung, Olga and Cohen-Or, Daniel},
  journal={Advances in Neural Information Processing Systems},
  volume={34},
  pages={8820--8832},
  year={2021}
}

@article{landgraf2022pins,
  title={Pins: progressive implicit networks for multi-scale neural representations},
  author={Landgraf, Zoe and Hornung, Alexander Sorkine and Cabral, Ricardo Silveira},
  journal={arXiv preprint arXiv:2202.04713},
  year={2022}
}

@inproceedings{lin2021barf,
  title={Barf: Bundle-adjusting neural radiance fields},
  author={Lin, Chen-Hsuan and Ma, Wei-Chiu and Torralba, Antonio and Lucey, Simon},
  booktitle={Proceedings of the IEEE/CVF international conference on computer vision},
  pages={5741--5751},
  year={2021}
}

@article{zhao2024galore,
  title={Galore: Memory-efficient llm training by gradient low-rank projection},
  author={Zhao, Jiawei and Zhang, Zhenyu and Chen, Beidi and Wang, Zhangyang and Anandkumar, Anima and Tian, Yuandong},
  journal={arXiv preprint arXiv:2403.03507},
  year={2024}
}

@article{robert2024ldadam,
  title={Ldadam: Adaptive optimization from low-dimensional gradient statistics},
  author={Robert, Thomas and Safaryan, Mher and Modoranu, Ionut-Vlad and Alistarh, Dan},
  journal={arXiv preprint arXiv:2410.16103},
  year={2024}
}

@article{hao2024flora,
  title={Flora: Low-rank adapters are secretly gradient compressors},
  author={Hao, Yongchang and Cao, Yanshuai and Mou, Lili},
  journal={arXiv preprint arXiv:2402.03293},
  year={2024}
}

@article{mahdavinia2025low,
  title={Low-rank momentum factorization for memory efficient training},
  author={Mahdavinia, Pouria and Mahdavi, Mehrdad},
  journal={arXiv preprint arXiv:2507.08091},
  year={2025}
}

@article{zhang2023adalora,
  title={Adalora: Adaptive budget allocation for parameter-efficient fine-tuning},
  author={Zhang, Qingru and Chen, Minshuo and Bukharin, Alexander and Karampatziakis, Nikos and He, Pengcheng and Cheng, Yu and Chen, Weizhu and Zhao, Tuo},
  journal={arXiv preprint arXiv:2303.10512},
  year={2023}
}

@article{li2018measuring,
  title={Measuring the intrinsic dimension of objective landscapes},
  author={Li, Chunyuan and Farkhoor, Heerad and Liu, Rosanne and Yosinski, Jason},
  journal={arXiv preprint arXiv:1804.08838},
  year={2018}
}

@article{gur2018gradient,
  title={Gradient descent happens in a tiny subspace},
  author={Gur-Ari, Guy and Roberts, Daniel A and Dyer, Ethan},
  journal={arXiv preprint arXiv:1812.04754},
  year={2018}
}

@inproceedings{tancik2021learned,
  title={Learned initializations for optimizing coordinate-based neural representations},
  author={Tancik, Matthew and Mildenhall, Ben and Wang, Terrance and Schmidt, Divi and Srinivasan, Pratul P and Barron, Jonathan T and Ng, Ren},
  booktitle={Proceedings of the IEEE/CVF conference on computer vision and pattern recognition},
  pages={2846--2855},
  year={2021}
}

@inproceedings{chen2021learning,
  title={Learning continuous image representation with local implicit image function},
  author={Chen, Yinbo and Liu, Sifei and Wang, Xiaolong},
  booktitle={Proceedings of the IEEE/CVF conference on computer vision and pattern recognition},
  pages={8628--8638},
  year={2021}
}

@article{xu2019frequency,
  title={Frequency principle: Fourier analysis sheds light on deep neural networks},
  author={Xu, Zhi-Qin John and Zhang, Yaoyu and Luo, Tao and Xiao, Yanyang and Ma, Zheng},
  journal={arXiv preprint arXiv:1901.06523},
  year={2019}
}

@inproceedings{gupta2018shampoo,
  title={Shampoo: Preconditioned stochastic tensor optimization},
  author={Gupta, Vineet and Koren, Tomer and Singer, Yoram},
  booktitle={International Conference on Machine Learning},
  pages={1842--1850},
  year={2018},
  organization={PMLR}
}

@article{vyas2024soap,
  title={Soap: Improving and stabilizing shampoo using adam},
  author={Vyas, Nikhil and Morwani, Depen and Zhao, Rosie and Kwun, Mujin and Shapira, Itai and Brandfonbrener, David and Janson, Lucas and Kakade, Sham},
  journal={arXiv preprint arXiv:2409.11321},
  year={2024}
}

@article{wang2025conda,
  title={Conda: Column-Normalized Adam for Training Large Language Models Faster},
  author={Wang, Junjie and Zhou, Pan and Dong, Yiming and Li, Huan and Li, Jia and Zhou, Xun and Lao, Qicheng and Fang, Cong and Lin, Zhouchen},
  journal={arXiv preprint arXiv:2509.24218},
  year={2025}
}

@article{xu2026width,
  title={On the Width Scaling of Neural Optimizers Under Matrix Operator Norms I: Row/Column Normalization and Hyperparameter Transfer},
  author={Xu, Ruihan and Li, Jiajin and Lu, Yiping},
  journal={arXiv preprint arXiv:2603.09952},
  year={2026}
}

@inproceedings{rezaeian2025sl2a,
  title={SL2A-INR: Single-Layer Learnable Activation for Implicit Neural Representation},
  author={Rezaeian, Reza and Heidari, Moein and Azad, Reza and Merhof, Dorit and Soltanian-Zadeh, Hamid and Hacihaliloglu, Ilker},
  booktitle={Proceedings of the IEEE/CVF International Conference on Computer Vision},
  pages={26065--26074},
  year={2025}
}

@article{choi2019empirical,
  title={On empirical comparisons of optimizers for deep learning},
  author={Choi, Dami and Shallue, Christopher J and Nado, Zachary and Lee, Jaehoon and Maddison, Chris J and Dahl, George E},
  journal={arXiv preprint arXiv:1910.05446},
  year={2019}
}

@inproceedings{akiba2019optuna,
  title={Optuna: A next-generation hyperparameter optimization framework},
  author={Akiba, Takuya and Sano, Shotaro and Yanase, Toshihiko and Ohta, Takeru and Koyama, Masanori},
  booktitle={Proceedings of the 25th ACM SIGKDD international conference on knowledge discovery \& data mining},
  pages={2623--2631},
  year={2019}
}

@inproceedings{wang2023neus2,
  title={Neus2: Fast learning of neural implicit surfaces for multi-view reconstruction},
  author={Wang, Yiming and Han, Qin and Habermann, Marc and Daniilidis, Kostas and Theobalt, Christian and Liu, Lingjie},
  booktitle={Proceedings of the IEEE/CVF international conference on computer vision},
  pages={3295--3306},
  year={2023}
}

@article{xu2025understanding,
  title={On understanding and overcoming spectral biases of deep neural network learning methods for solving PDEs},
  author={Xu, Zhi-Qin John and Zhang, Lulu and Cai, Wei},
  journal={Journal of Computational Physics},
  volume={530},
  pages={113905},
  year={2025},
  publisher={Elsevier}
}

@article{schotthofer2025geometric,
  title={A geometric framework for momentum-based optimizers for low-rank training},
  author={Schotth{\"o}fer, Steffen and Klein, Timon and Kusch, Jonas},
  journal={arXiv preprint arXiv:2506.17475},
  year={2025}
}

@article{wang2026taming,
  title={Taming Momentum: Rethinking Optimizer States Through Low-Rank Approximation},
  author={Wang, Zhengbo and Liang, Jian and He, Ran and Wang, Zilei and Tan, Tieniu},
  journal={arXiv preprint arXiv:2602.24283},
  year={2026}
}

@article{yu2025altlora,
  title={Altlora: Towards better gradient approximation in low-rank adaptation with alternating projections},
  author={Yu, Xin and Wang, Yujia and Chen, Jinghui and Xue, Lingzhou},
  journal={arXiv preprint arXiv:2505.12455},
  year={2025}
}

@article{schotthofer2025dynamical,
  title={Dynamical low-rank compression of neural networks with robustness under adversarial attacks},
  author={Schotth{\"o}fer, Steffen and Yang, H Lexie and Schnake, Stefan},
  journal={arXiv preprint arXiv:2505.08022},
  year={2025}
}

@article{pang2026htmuon,
  title={HTMuon: Improving Muon via Heavy-Tailed Spectral Correction},
  author={Pang, Tianyu and Fang, Yujie and Liu, Zihang and Deng, Shenyang and Hsiung, Lei and Yu, Shuhua and Yang, Yaoqing},
  journal={arXiv preprint arXiv:2603.10067},
  year={2026}
}

@book{doi:10.1137/1.9780898717778,
author = {Higham, Nicholas J.},
title = {Functions of Matrices},
publisher = {Society for Industrial and Applied Mathematics},
year = {2008},
doi = {10.1137/1.9780898717778},
address = {},
edition   = {},
URL = {https://epubs.siam.org/doi/abs/10.1137/1.9780898717778},
eprint = {https://epubs.siam.org/doi/pdf/10.1137/1.9780898717778}
}

@article{mildenhall2021nerf,
  title={Nerf: Representing scenes as neural radiance fields for view synthesis},
  author={Mildenhall, Ben and Srinivasan, Pratul P and Tancik, Matthew and Barron, Jonathan T and Ramamoorthi, Ravi and Ng, Ren},
  journal={Communications of the ACM},
  volume={65},
  number={1},
  pages={99--106},
  year={2021},
  publisher={ACM New York, NY, USA}
}

@inproceedings{bergstra2011algorithms,
  title={Algorithms for Hyper-Parameter Optimization},
  author={Bergstra, James and Bardenet, Remi and Bengio, Yoshua and Kegl, Balazs},
  booktitle={Advances in Neural Information Processing Systems},
  volume={24},
  pages={2546--2554},
  year={2011}
}

@article{kingma2014adam,
  title={Adam: A method for stochastic optimization},
  author={Kingma, Diederik P and Ba, Jimmy},
  journal={arXiv preprint arXiv:1412.6980},
  year={2014}
}

@article{muller2022instant,
  title={Instant neural graphics primitives with a multiresolution hash encoding},
  author={M{\"u}ller, Thomas and Evans, Alex and Schied, Christoph and Keller, Alexander},
  journal={ACM transactions on graphics (TOG)},
  volume={41},
  number={4},
  pages={1--15},
  year={2022},
  publisher={ACM New York, NY, USA}
}

@inproceedings{barron2021mip,
  title={Mip-nerf: A multiscale representation for anti-aliasing neural radiance fields},
  author={Barron, Jonathan T and Mildenhall, Ben and Tancik, Matthew and Hedman, Peter and Martin-Brualla, Ricardo and Srinivasan, Pratul P},
  booktitle={Proceedings of the IEEE/CVF international conference on computer vision},
  pages={5855--5864},
  year={2021}
}

@misc{jordan2024muon,
  author       = {Keller Jordan and Yuchen Jin and Vlado Boza and Jiacheng You and
                  Franz Cesista and Laker Newhouse and Jeremy Bernstein},
  title        = {Muon: An optimizer for hidden layers in neural networks},
  year         = {2024},
  url          = {https://kellerjordan.github.io/posts/muon/}
}

@article{loshchilov2017decoupled,
  title={Decoupled weight decay regularization},
  author={Loshchilov, Ilya and Hutter, Frank},
  journal={arXiv preprint arXiv:1711.05101},
  year={2017}
}

@InProceedings{Timofte_2017_CVPR_Workshops,
author = {Timofte, Radu and Agustsson, Eirikur and Van Gool, Luc and Yang, Ming-Hsuan and Zhang, Lei and Lim, Bee and others},
title = {NTIRE 2017 Challenge on Single Image Super-Resolution: Methods and Results},
booktitle = {The IEEE Conference on Computer Vision and Pattern Recognition (CVPR) Workshops},
month = {July},
year = {2017}
}

@misc{kodak,
  author       = {Rich Franzen},
  title        = {Kodak Lossless True Color Image Suite},
  howpublished = {\url{http://r0k.us/graphics/kodak/}},
  year         = {1999},
  note         = {Online; accessed 5 May 2026}
}

@article{mildenhall2019local,
  title={Local light field fusion: Practical view synthesis with prescriptive sampling guidelines},
  author={Mildenhall, Ben and Srinivasan, Pratul P and Ortiz-Cayon, Rodrigo and Kalantari, Nima Khademi and Ramamoorthi, Ravi and Ng, Ren and Kar, Abhishek},
  journal={ACM Transactions on Graphics (TOG)},
  volume={38},
  number={4},
  pages={1--14},
  year={2019},
  publisher={ACM New York, NY, USA}
}

@INPROCEEDINGS{roy2007effective,
  author={Roy, Olivier and Vetterli, Martin},
  booktitle={2007 15th European Signal Processing Conference}, 
  title={The effective rank: A measure of effective dimensionality}, 
  year={2007},
  volume={},
  number={},
  pages={606-610},
  doi={}}

@article{he2025low,
  title={Low-rank orthogonalization for large-scale matrix optimization with applications to foundation model training},
  author={He, Chuan and Deng, Zhanwang and Lu, Zhaosong},
  journal={arXiv preprint arXiv:2509.11983},
  year={2025}
}

@article{lazzari2023understanding,
  title={Understanding the spectral bias of coordinate based MLPs via training dynamics},
  author={Lazzari, John and Liu, Xiuwen},
  journal={arXiv preprint arXiv:2301.05816},
  year={2023}
}
}
\newpage
\appendix
\raggedbottom

\makeatletter
\@addtoreset{table}{section}
\@addtoreset{figure}{section}
\makeatother

\renewcommand{\thetable}{\Alph{section}.\arabic{table}}
\renewcommand{\thefigure}{\Alph{section}.\arabic{figure}}

\section{Additional Experiments}
\label{app:additional}
\vspace{-0.05in}
\subsection{Unified evaluation across architectures and tasks}
\label{app:unified_eval}
\vspace{-0.06in}
We compare Adam, Muon, and ProPolar on ReLU MLP, SIREN, WIRE, and FINER under identical settings across optimizers. All experiments in this section use a single NVIDIA RTX PRO 6000 GPU, and absolute numbers differ from those in the main paper.
\vspace{-0.1in}
\paragraph{Image overfitting.}
Table~\ref{tab:a_app_image_overfit} reports PSNR on Kodak24. ProPolar achieves the highest mean PSNR on all four architectures.
\vspace{-0.1in}
\paragraph{SDF fitting.}
Table~\ref{tab:a_app_sdf} reports two errors on Armadillo and Dragon for WIRE and FINER. SDF denotes the mean absolute error between the predicted and ground-truth signed distances over all query points, and Near-surf.\ denotes the same error over query points within a distance of $\epsilon$ from the surface. ProPolar attains the lowest SDF error in three of the four settings, while Near-surf.\ shows no consistent winner. This result suggests that the gain of ProPolar on SDF fitting comes from reducing errors away from the surface.
\vspace{-0.13in}
\begin{table}[h]
\centering
\footnotesize
\setlength{\tabcolsep}{8pt}%
\renewcommand{\arraystretch}{0.95}%
\setlength{\aboverulesep}{1.2pt}%
\setlength{\belowrulesep}{1.2pt}%
\caption{\textbf{Image overfitting on Kodak24.} PSNR as mean $\pm$ std over 24 images. Best per architecture in bold.}
\label{tab:a_app_image_overfit}
\providecommand{\mstd}[2]{#1\,$\pm$\,#2}%
\providecommand{\mstdbf}[2]{\textbf{#1}\,{\boldmath$\pm$}\,\textbf{#2}}%
\begin{tabular}{@{}lccc}
\toprule[0.8pt]
\multirow{2}{*}{\textbf{Model}} & \multicolumn{3}{c}{\textbf{PSNR}\,$\uparrow$} \\
\cmidrule(lr){2-4}
& Adam~\cite{kingma2014adam} & Muon~\cite{jordan2024muon} & \bestcell{\textbf{Ours}} \\
\midrule
ReLU MLP & \mstd{22.53}{2.85} & \mstd{24.92}{2.91} & \bestcell{\mstdbf{27.33}{5.38}} \\
SIREN    & \mstd{37.68}{2.29} & \mstd{40.86}{2.29} & \bestcell{\mstdbf{41.22}{3.51}} \\
WIRE     & \mstd{29.41}{1.61} & \mstd{36.23}{2.73} & \bestcell{\mstdbf{36.51}{2.89}} \\
FINER    & \mstd{38.16}{1.98} & \mstd{42.27}{3.03} & \bestcell{\mstdbf{43.05}{3.02}} \\
\bottomrule[0.8pt]
\end{tabular}
\vspace{-0.12in}
\end{table}

\vspace{-0.13in}
\begin{table}[h]
\centering
\footnotesize
\setlength{\tabcolsep}{7pt}%
\renewcommand{\arraystretch}{0.9}%
\setlength{\aboverulesep}{1.2pt}%
\setlength{\belowrulesep}{1.2pt}%
\caption{\textbf{SDF fitting on Armadillo and Dragon.} Mean absolute error over all query points and near-surface points. Best per architecture and shape in bold.}
\vspace{2pt}
\label{tab:a_app_sdf}
\begin{tabular}{@{}llcccc@{}}
\toprule[0.8pt]
 & & \multicolumn{2}{c}{\textbf{Armadillo}} & \multicolumn{2}{c}{\textbf{Dragon}} \\
\cmidrule(lr){3-4} \cmidrule(l){5-6}
\textbf{Model} & \textbf{Optimizer} & SDF $\downarrow$ & Near-surf. $\downarrow$ & SDF $\downarrow$ & Near-surf. $\downarrow$ \\
\midrule
\multirow{3}{*}{WIRE}  & Adam & 0.03917          & \textbf{0.00092} & 0.03186          & 0.02890          \\
                       & Muon & 0.02181          & 0.00249          & \textbf{0.03157} & \textbf{0.02861} \\
                       & \bestcell{Ours} & \bestcell{\textbf{0.02174}} & \bestcell{0.00155} & \bestcell{0.03175} & \bestcell{0.02869} \\
\cmidrule(l){2-6}
\multirow{3}{*}{FINER} & Adam & 0.08490          & \textbf{0.00558} & 0.04006          & 0.00537          \\
                       & Muon & 0.08070          & 0.00562          & 0.03761          & \textbf{0.00511} \\
                       & \bestcell{Ours} & \bestcell{\textbf{0.07580}} & \bestcell{0.00587} & \bestcell{\textbf{0.03688}} & \bestcell{0.00556} \\
\bottomrule[0.8pt]
\end{tabular}
\vspace{-0.12in}
\end{table}

\subsection{Additional NeRF Results at 200K Iterations}
\label{sec:nerf_200k_results}
\vspace{-0.06in}
\noindent
This appendix provides additional quantitative results for vanilla NeRF trained for 200K iterations in Table~\ref{tab:nerf_quan_200k}.

\providecommand{\avgcell}[1]{\cellcolor{gray!12}#1}
\providecommand{\avgcellbold}[1]{\cellcolor{gray!12}\textbf{#1}}
\providecommand{\bestcell}[1]{#1}

\noindent
\begin{minipage}{\linewidth}
\small
\captionsetup{
    width=\linewidth,
    margin=0pt,
    justification=raggedright,
    singlelinecheck=false,
    format=plain,
    position=top,
    skip=2pt
}

\captionof{table}{\small\textbf{Additional 200K quantitative comparison on vanilla NeRF~\cite{mildenhall2021nerf}.}
Scene-level PSNR and SSIM are reported on the LLFF and Synthetic NeRF datasets.
The best result for each scene is highlighted in bold.}
\label{tab:nerf_quan_200k}

\vspace{2pt}

\centering

{%
\renewcommand{\arraystretch}{1.12}%
\setlength{\tabcolsep}{2.2pt}%
\setlength{\aboverulesep}{0.15pt}%
\setlength{\belowrulesep}{0.15pt}%
\setlength{\cmidrulesep}{0.15pt}%

\resizebox{0.83\linewidth}{!}{%
\begin{tabular}{@{}cc*{8}{c}c@{}}
\toprule
\multirow{2}{*}{\textbf{Metric}} & \multirow{2}{*}{\textbf{Method}}
& \multicolumn{8}{c}{\textbf{LLFF}}
& \\
\cmidrule(lr){3-10}
& & Fern & Flower & Fortress & Horns & Leaves & Orchids & Room & T-Rex
& \avgcell{\textcolor{gray}{Avg.}} \\
\midrule

\multirow{3}{*}{PSNR $\uparrow$}
& Adam~\cite{kingma2014adam}
& 26.91 & 28.25 & 32.55 & 28.87 & 22.43 & 21.34 & 32.76 & 27.83
& \avgcell{27.62} \\
& Muon~\cite{jordan2024muon}
& \underline{27.20} & \underline{28.86} & \underline{33.03} & \underline{29.57} & \underline{22.62} & \underline{21.57} & \underline{34.17} & \underline{28.46}
& \avgcell{\underline{28.18}} \\
& \bestcell{Ours}
& \bestcell{\textbf{27.42}} & \bestcell{\textbf{28.96}} & \bestcell{\textbf{33.17}} & \bestcell{\textbf{29.66}}
& \bestcell{\textbf{22.72}} & \bestcell{\textbf{21.72}} & \bestcell{\textbf{34.21}} & \bestcell{\textbf{28.92}}
& \avgcellbold{\textbf{28.35}} \\

\midrule

\multirow{3}{*}{SSIM $\uparrow$}
& Adam~\cite{kingma2014adam}
& 0.8588 & 0.8879 & 0.9235 & 0.8978 & \underline{0.8179} & 0.7509 & 0.9556 & 0.9208
& \avgcell{0.8767} \\
& Muon~\cite{jordan2024muon}
& \underline{0.8665} & \underline{0.8969} & \underline{0.9332} & \textbf{0.9172} & 0.8135 & \underline{0.7546} & \textbf{0.9666} & \underline{0.9342}
& \avgcell{\underline{0.8853}} \\
& \bestcell{Ours}
& \bestcell{\textbf{0.8746}} & \bestcell{\textbf{0.9058}} & \bestcell{\textbf{0.9357}} & \bestcell{\underline{0.9163}}
& \bestcell{\textbf{0.8308}} & \bestcell{\textbf{0.7733}} & \bestcell{\underline{0.9649}} & \bestcell{\textbf{0.9368}}
& \avgcellbold{\textbf{0.8923}} \\
\bottomrule
\end{tabular}%
}

\vspace{1pt}

\resizebox{0.83\linewidth}{!}{%
\begin{tabular}{@{}cc*{8}{c}c@{}}
\toprule
\multirow{2}{*}{\textbf{Metric}} & \multirow{2}{*}{\textbf{Method}}
& \multicolumn{8}{c}{\textbf{Synthetic}}
& \\
\cmidrule(lr){3-10}
& & Chair & Drums & Ficus & Hotdog & Lego & Materials & Mic & Ship
& \avgcell{\textcolor{gray}{Avg.}} \\
\midrule

\multirow{3}{*}{PSNR $\uparrow$}
& Adam~\cite{kingma2014adam}
& 33.97 & {25.55} & {28.99} & 36.69 & {31.65} & 29.35 & 33.18 & 29.22
& \avgcell{31.08} \\
& Muon~\cite{jordan2024muon}
& \underline{34.30} & \underline{25.67} & \underline{29.19} & \underline{37.20} & \underline{31.93} & \underline{29.85} & \underline{33.46} & \underline{29.49}
& \avgcell{\underline{31.39}} \\
& \bestcell{Ours}
& \bestcell{\textbf{35.04}} & \bestcell{\textbf{25.93}} & \bestcell{\textbf{29.80}} & \bestcell{\textbf{37.90}}
& \bestcell{\textbf{32.61}} & \bestcell{\textbf{30.34}} & \bestcell{\textbf{34.07}} & \bestcell{\textbf{29.81}}
& \avgcellbold{\textbf{31.94}} \\

\midrule

\multirow{3}{*}{SSIM $\uparrow$}
& Adam~\cite{kingma2014adam}
& 0.9775 & {0.9295} & {0.9655} & 0.9791 & 0.9651 & 0.9578 & 0.9785 & 0.8717
& \avgcell{0.9531} \\
& Muon~\cite{jordan2024muon}
& \underline{0.9790} & \underline{0.9320} & \underline{0.9658} & \underline{0.9823} & \underline{0.9686} & \underline{0.9623} & \underline{0.9812} & \underline{0.8842}
& \avgcell{\underline{0.9541}} \\
& \bestcell{Ours}
& \bestcell{\textbf{0.9826}} & \bestcell{\textbf{0.9346}} & \bestcell{\textbf{0.9707}} & \bestcell{\textbf{0.9839}}
& \bestcell{\textbf{0.9729}} & \bestcell{\textbf{0.9660}} & \bestcell{\textbf{0.9826}} & \bestcell{\textbf{0.8846}}
& \avgcellbold{\textbf{0.9597}} \\
\bottomrule
\end{tabular}%
}

}%

\end{minipage}

\section{Schedule Hyperparameters}
\label{app:schedule_hparams}
\begin{wraptable}[6]{r}{0.55\textwidth}
\vspace{-30pt}
\centering
\small
\caption{\textbf{Rank schedule hyperparameters}, shared across all experiments.}
\label{tab:app_schedule_hparams}
\begin{tabular}{@{}lll@{}}
\toprule[0.8pt]
\textbf{Symbol} & \textbf{Value} & \textbf{Role} \\
\midrule
$\rho$                  & $0.90$       & Energy threshold for rank initialization \\
$T_0$                   & $8$          & Initialization phase length \\
$T_g$                   & $0.8T - T_0$ & Rank-growth phase length \\
$\Delta_{\mathrm{sat}}$ & $0$          & Delay before LR decay \\
\bottomrule[0.8pt]
\end{tabular}
\end{wraptable}

We set $T_g = 0.8T - T_0$ so that the rank saturates at 80\% of training for any budget $T$, after which the learning rate decays linearly ($\Delta_{\mathrm{sat}} = 0$). This ensures the rank schedule finishes before the learning rate anneals. All values were fixed a priori without per-experiment tuning.

\section{Rank Schedule Ablation over Training}
\label{app:schedule}
\vspace{-0.03in}
We extend the rank schedule ablation in Table~\ref{tab:fig6_evidence} by tracking PSNR during training. The increasing-rank schedule attains the highest mean PSNR at every checkpoint from 1,000 iterations onward. The margin over the next-best schedule stays at or below 0.22\,dB until 2,000 iterations and grows to 1.80\,dB at 3,000 iterations, which is consistent with the late increase in effective rank in Figure~\ref{fig:rank_plot}.
\vspace{-0.14in}

\begin{table}[h]
\centering
\footnotesize
\setlength{\aboverulesep}{1.2pt}%
\setlength{\belowrulesep}{1.2pt}%
\caption{\textbf{PSNR on Kodak24 at intermediate checkpoints.} Mean $\pm$ std over 24 images, measured every 500 iterations on an NVIDIA RTX PRO 6000 GPU. All variants follow the setting of Table~\ref{tab:fig6_evidence} except the rank trajectory. Best mean at each checkpoint in bold.}
\vspace{4pt}
\label{tab:app_schedule_ckpt}
\newcommand{\mstd}[2]{#1\,$\pm$\,#2}%
\newcommand{\mstdbf}[2]{\textbf{#1}\,{\boldmath$\pm$}\,\textbf{#2}}%
\setlength{\tabcolsep}{3pt}%
\begin{tabular}{@{}lccccccc@{}}
\toprule[0.8pt]
\multirow{2}{*}{\textbf{Method}} & \multicolumn{7}{c}{\textbf{Iteration}} \\
\cmidrule(l){2-8}
 & 0 & 500 & 1,000 & 1,500 & 2,000 & 2,500 & 3,000 \\
\midrule
Decreasing rank & \mstd{7.34}{1.77} & \mstd{22.16}{2.31}   & \mstd{22.69}{2.80}   & \mstd{21.13}{7.97}   & \mstd{21.37}{5.34}   & \mstd{22.96}{4.33}   & \mstd{25.53}{6.45}   \\
Full rank       & \mstd{7.34}{1.77} & \mstd{19.60}{2.22}   & \mstd{20.48}{5.08}   & \mstd{20.93}{4.62}   & \mstd{21.57}{5.00}   & \mstd{22.50}{5.38}   & \mstd{25.22}{7.08}   \\
Fixed rank      & \mstd{7.34}{1.77} & \mstdbf{22.19}{2.26} & \mstd{22.44}{3.32}   & \mstd{22.51}{3.21}   & \mstd{22.83}{3.18}   & \mstd{22.70}{5.33}   & \mstd{25.25}{5.13}   \\
Increasing rank & \mstd{7.34}{1.77} & \mstd{22.14}{2.30}   & \mstdbf{22.78}{2.28} & \mstdbf{22.64}{3.61} & \mstdbf{23.05}{3.70} & \mstdbf{24.36}{4.23} & \mstdbf{27.33}{5.37} \\
\bottomrule[0.8pt]
\end{tabular}
\end{table}

\vspace{-1pt}

\section{Hyperparameter Optimization Details}
\label{sec:hpo_details}
\vspace{-0.02in}
\paragraph{TPE-based HPO Details.}
We use Optuna's Tree-structured Parzen Estimator (TPE) for the NeRF HPO comparison.
Each sampled configuration is trained for 100K iterations and evaluated by validation-set mean PSNR.
All optimizers are tuned with 20 trials under the same search budget, and the final configuration is selected based on validation PSNR.
\vspace{-0.08in}
\begin{table}[H]
\centering
\vspace{-6pt}
\captionsetup{
    width=\linewidth,
    margin=0pt,
    justification=raggedright,
    singlelinecheck=false,
    format=plain,
    position=top,
    skip=3pt
}
\caption{\textbf{TPE search spaces for NeRF HPO.} Each optimizer uses 20 trials with 100K training iterations per trial, and the best configuration is selected by validation PSNR.}
\label{tab:hpo_search_space}
\footnotesize
\renewcommand{\arraystretch}{1.0}%
\setlength{\tabcolsep}{10pt}%
\setlength{\aboverulesep}{1.2pt}%
\setlength{\belowrulesep}{1.2pt}%
\begin{tabular}{@{}llll@{}}
\toprule
\textbf{Method} & \textbf{Tuned hyperparameters} & \textbf{Search space} & \textbf{Sampling} \\
\midrule
Adam
& \makecell[l]{Adam learning rate ($\eta_{\mathrm{Adam}}$) \\ Learning-rate decay}
& \makecell[l]{$\eta_{\mathrm{Adam}} \in [3\mathrm{e}{-4},\,3\mathrm{e}{-3}]$ \\ Decay $\in [100,\,500]$}
& \makecell[l]{log-uniform \\ integer uniform} \\
\midrule
Muon
& \makecell[l]{Adam branch learning rate ($\eta_{\mathrm{Adam}}$) \\ Learning-rate decay \\ Muon learning rate ($\eta_{\mathrm{Muon}}$)}
& \makecell[l]{$\eta_{\mathrm{Adam}} \in [3\mathrm{e}{-4},\,3\mathrm{e}{-3}]$ \\ Decay $\in [100,\,500]$ \\ $\eta_{\mathrm{Muon}} \in [3\mathrm{e}{-4},\,5\mathrm{e}{-3}]$}
& \makecell[l]{log-uniform \\ integer uniform \\ log-uniform} \\
\midrule
Ours
& \makecell[l]{Adam branch learning rate ($\eta_{\mathrm{Adam}}$) \\ Learning-rate decay \\ Projected-polar learning rate ($\eta_{\mathrm{polar}}$)}
& \makecell[l]{$\eta_{\mathrm{Adam}} \in [3\mathrm{e}{-4},\,3\mathrm{e}{-3}]$ \\ Decay $\in [100,\,500]$ \\ $\eta_{\mathrm{polar}} \in [3\mathrm{e}{-4},\,5\mathrm{e}{-3}]$}
& \makecell[l]{log-uniform \\ integer uniform \\ log-uniform} \\
\bottomrule
\end{tabular}
\end{table}

\label{sec:tpe_hpo_details}
\vspace{-0.2in}

\section{Additional Applications}
\label{sec:fine}

\subsection{Model Fine-tuning}
\begin{figure}[H]
\centering
\begin{tabular}{@{}c@{\hspace{-2pt}}c@{}}
\LegoZoom{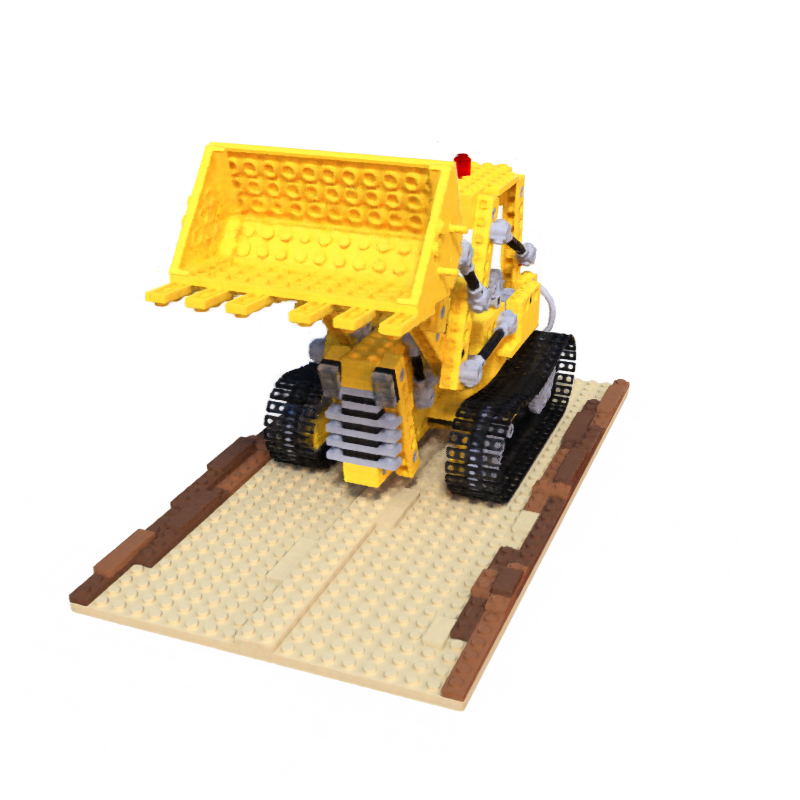}{0.5\linewidth}{Muon} &
\LegoZoom{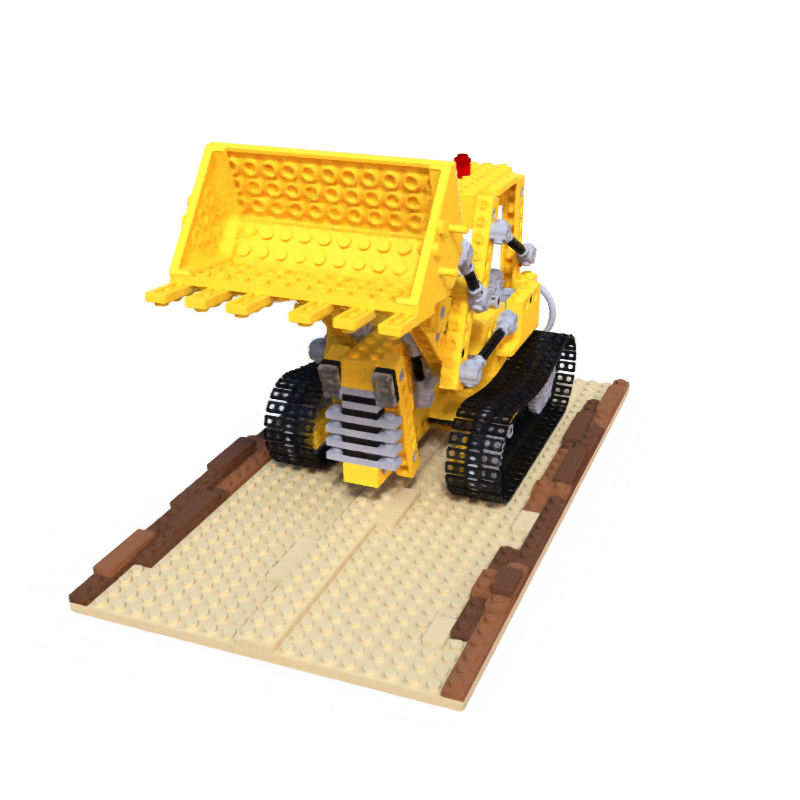}{0.5\linewidth}{Ours}
\end{tabular}
\caption{\textbf{Fine-tuning comparison on the Lego scene.}
Zoomed crops highlight brick boundaries during refinement.
The proposed optimizer recovers sharper details.}
\label{fig:nerf_finetune}
\end{figure}

Figure~\ref{fig:nerf_finetune} examines NeRF fine-tuning on the Lego scene after the coarse reconstruction stage. 
The comparison targets the refinement stage, where optimizer updates must recover localized high-frequency residuals from an already recovered coarse structure. 
The matched crop centers on brick boundaries, which serve as a sensitive probe of local detail recovery. 
Muon yields smoother boundary transitions in the magnified region. 
The proposed optimizer preserves sharper edges and more regular brick contours. 
The proposed method preserves finer details more clearly than the baseline, even at a lower iteration count.

\subsection{Image denoising}
\label{app:denoising}

Image denoising penalizes fitting high-frequency content, and we use it to test whether ProPolar overfits noise. Each INR fits an image corrupted by Poisson noise at $\tau = 30$ and Gaussian readout noise at an SNR of 2\,dB, and PSNR is measured against the clean image. All methods share the architecture, iteration budget, and weight decay for each INR, and the full settings are provided in the released code.

Table~\ref{tab:app_denoising} shows that our method attains the highest PSNR on all four architectures. The gain over the best baseline ranges from 0.20\,dB on WIRE to 1.20\,dB on FINER and exceeds the seed-to-seed standard deviation of each architecture. This result suggests that the benefit of our method is not limited to tasks that reward fitting high-frequency detail.

\vspace{-0.08in}

\begin{table}[h]
\centering
\footnotesize
\setlength{\tabcolsep}{8pt}%
\renewcommand{\arraystretch}{0.95}%
\setlength{\aboverulesep}{1.2pt}%
\setlength{\belowrulesep}{1.2pt}%
\caption{\textbf{Image denoising.} PSNR against the clean image as mean $\pm$ std over 10 random seeds. Best per architecture in bold.}
\label{tab:app_denoising}
\providecommand{\mstd}[2]{#1\,$\pm$\,#2}%
\providecommand{\mstdbf}[2]{\textbf{#1}\,{\boldmath$\pm$}\,\textbf{#2}}%
\begin{tabular}{@{}lccc}
\toprule[0.8pt]
\multirow{2}{*}{\textbf{Model}} & \multicolumn{3}{c}{\textbf{PSNR}\,$\uparrow$} \\
\cmidrule(lr){2-4}
& Adam~\cite{kingma2014adam} & Muon~\cite{jordan2024muon} & \bestcell{\textbf{Ours}} \\
\midrule
ReLU MLP & \mstd{22.04}{0.21} & \mstd{26.97}{0.12} & \bestcell{\mstdbf{28.09}{0.09}} \\
SIREN    & \mstd{25.87}{0.05} & \mstd{27.06}{0.05} & \bestcell{\mstdbf{27.79}{0.04}} \\
WIRE     & \mstd{28.56}{0.06} & \mstd{28.46}{0.05} & \bestcell{\mstdbf{28.76}{0.05}} \\
FINER    & \mstd{28.24}{0.06} & \mstd{28.62}{0.04} & \bestcell{\mstdbf{29.82}{0.05}} \\
\bottomrule[0.8pt]
\end{tabular}
\vspace{-0.12in}
\end{table}

\newpage

\newtcblisting{pycodebox}{
    colback=lightgray!10,
    colframe=gray!50,
    boxrule=0.5pt,
    arc=2pt,
    left=3pt,
    right=3pt,
    top=3pt,
    bottom=3pt,
    listing only,
    listing engine=listings,
    listing options={
        language=Python,
        basicstyle=\footnotesize\ttfamily,
        keywordstyle=\color{blue}\bfseries,
        commentstyle=\color{green!60!black},
        stringstyle=\color{red},
        showstringspaces=false,
        tabsize=4,
        breaklines=true,
        columns=flexible
    }
}

\section{Implementation of Progressive Rank and Rank-Aware Warmup-Stable-Decay Scheduler}
\label{appendix:rank-aware-wsd}

\begin{algorithm}[H]
    \caption{Code: Progressively Increasing Rank}
    \centering

\begin{pycodebox}
def cosine_rank(step, rank_start, rank_end, rank_warmup):
    if rank_warmup <= 1:
        return rank_end
    if step <= 1:
        return rank_start
    if step >= rank_warmup:
        return rank_end

    tau = (step - 1) / (rank_warmup - 1)
    tau = 0.5 * (1 - cos(pi * tau))
    rank = rank_start + tau * (rank_end - rank_start)
    return round(rank)


def update_progressive_rank(group, grad=None):
    group.step += 1

    # Optional: estimate initial rank from early gradient energy.
    if group.auto_rank_start and group.step <= group.probe_steps:
        group.rank_start = EstimateRankStart(grad)

    rank = cosine_rank(
        group.step,
        group.rank_start,
        group.rank_end,
        group.rank_warmup
    )

    group.current_rank = max(group.rank_min, rank)
    group.target_rank = max(group.rank_min, group.rank_end)

    group.rank_saturated = group.current_rank >= group.target_rank
    group.rank_progress = Progress(
        group.current_rank,
        group.rank_start,
        group.target_rank
    )

    return group.current_rank
\end{pycodebox}
\end{algorithm}

\begin{algorithm}[H]
    \caption{Code: Rank-Aware Warmup-Stable-Decay Scheduler}
    \centering

\begin{pycodebox}
def lr_ratio(step, warmup, decay_start, decay_end, min_ratio):
    if warmup > 0 and step < warmup:
        return (step + 1) / warmup

    if decay_start is None or step < decay_start:
        return 1.0

    if step >= decay_end:
        return min_ratio

    tau = (step - decay_start) / max(1, decay_end - decay_start)
    return 1.0 - (1.0 - min_ratio) * tau


def rank_aware_wsd(step, optimizer):
    status = ReadRankMetadata(optimizer)

    # Progressive rank: keep peak LR until rank saturation.
    if decay_window_is_unresolved():
        if status.has_progressive_rank:
            if status.rank_saturated:
                decay_start = step + patience
                decay_end = total_steps - 1
            else:
                decay_start = None
                decay_end = None

        # Full-rank or fixed-rank Muon does not wait for late decay.
        elif status.is_full_rank_or_fixed_rank:
            decay_start = round(full_start_frac * total_steps)
            decay_end = round(full_end_frac * total_steps)

        else:
            decay_start = round(default_start_frac * total_steps)
            decay_end = round(default_end_frac * total_steps)

    ratio = lr_ratio(
        step,
        warmup_steps,
        decay_start,
        decay_end,
        min_lr_ratio
    )

    lr_adam = base_lr_adam * ratio
    lr_muon = base_lr_muon * ratio

    return lr_adam, lr_muon, status
\end{pycodebox}
\end{algorithm}

\begin{algorithm}[H]
    \caption{Code: Progressive Rank Optimizer with Rank-Aware WSD}
    \centering

\begin{pycodebox}
for step in range(total_steps):
    loss = model(batch)
    loss.backward()

    # Scheduler reads rank metadata from optimizer groups.
    lr_adam, lr_muon, status = rank_aware_wsd(step, optimizer)
    SetLearningRate(optimizer, lr_adam, lr_muon)

    for group in optimizer.muon_groups:
        rank = update_progressive_rank(group, grad=group.grad)
        update = LowRankMuonUpdate(group.grad, rank)
        group.params -= lr_muon * update

    for group in optimizer.adam_groups:
        update = AdamUpdate(group.grad)
        group.params -= lr_adam * update

    optimizer.zero_grad()
\end{pycodebox}
\end{algorithm}

\end{document}